\documentclass[nonacm]{aamas}

\usepackage{balance} % for balancing columns on the final page
\usepackage{lineno}
\usepackage{comment}
\usepackage{enumitem}
\usepackage{graphicx}
\usepackage{wrapfig}
\usepackage{amsmath}
\usepackage{algorithm}
\usepackage{algpseudocode}
\usepackage{csquotes}
\usepackage[dvipsnames]{xcolor}
\setcopyright{ifaamas}
\acmConference[AAMAS '26]{Proc.\@ of the 25th International Conference
on Autonomous Agents and Multiagent Systems (AAMAS 2026)}{May 25 -- 29, 2026}
{Paphos, Cyprus}{C.~Amato, L.~Dennis, V.~Mascardi, J.~Thangarajah (eds.)}
\copyrightyear{2026}
\acmYear{2026}
\acmDOI{}
\acmPrice{}
\acmISBN{}
\title[AAMAS-2026 Formatting Instructions]{DiffVAS: Diffusion-Guided Visual Active Search in Partially Observable Environments}

\author{Anindya Sarkar$^*$}\thanks{* Equal contribution.}
\orcid{0000-0001-8037-9070}
\affiliation{
  \institution{Washington University in St. Louis}
  \city{St. Louis}
  \country{United States}}
\email{anindyasarkar.ece@gmail.com}

\author{Srikumar Sastry$^*$}
\orcid{0000-0002-4646-9416}
\affiliation{
  \institution{Washington University in St. Louis}
  \city{St. Louis}
  \country{United States}}
\email{s.sastry@wustl.edu}

\author{Aleksis Pirinen}
\orcid{0000-0002-5299-142X}
\affiliation{
  \institution{RISE Research Institutes of Sweden}
  \institution{Climate AI Nordics}
  \city{Lund}
  \country{Sweden}}
\email{aleksis.pirinen@ri.se}

\author{Nathan Jacobs}
\orcid{0000-0002-4242-8967}
\affiliation{
  \institution{Washington University in St. Louis}
  \city{St. Louis}
  \country{United States}}
\email{jacobsn@wustl.edu}

\author{Yevgeniy Vorobeychik}
\orcid{0000-0003-2471-5345}
\affiliation{
  \institution{Washington University in St. Louis}
  \city{St. Louis}
  \country{United States}}
\email{yvorobeychik@wustl.edu}

\begin{abstract}
Visual active search (VAS) has been introduced as a modeling framework that leverages visual cues to direct aerial (e.g.,~UAV-based) exploration and pinpoint areas of interest within extensive geospatial regions. Potential applications of VAS include detecting hotspots for rare wildlife poaching, aiding search-and-rescue missions, and uncovering illegal trafficking of weapons, among other uses. Previous VAS approaches assume that the entire search space is known upfront, which is often unrealistic due to constraints such as a restricted field of view and high acquisition costs, and they typically learn policies tailored to specific target objects, which limits their ability to search for multiple target categories simultaneously. In this work, we propose \emph{DiffVAS}, a target-conditioned policy that searches for diverse objects simultaneously according to task requirements in partially observable environments, which advances the deployment of visual active search policies in real-world applications. DiffVAS leverages a diffusion model to reconstruct the entire geospatial area from sequentially observed partial glimpses, which enables a target-conditioned reinforcement learning-based planning module to effectively reason and guide subsequent search steps. Extensive experiments demonstrate that DiffVAS excels in searching diverse objects in partially observable environments, significantly surpassing state-of-the-art methods on several datasets. Code and models are available at this \href{https://github.com/mvrl/Multi-objective-active-sampling-for-inpainting}{\textcolor{blue}{link}}. 
\end{abstract}


\newcommand{\BibTeX}{\rm B\kern-.05em{\sc i\kern-.025em b}\kern-.08em\TeX}

\begin{document}

%%% The following commands remove the headers in your paper. For final 
%%% papers, these will be inserted during the pagination process.

\pagestyle{fancy}
\fancyhead{}

%%% The next command prints the information defined in the preamble.

\maketitle 

%%%%%%%%%%%%%%%%%%%%%%%%%%%%%%%%%%%%%%%%%%%%%%%%%%%%%%%%%%%%%%%%%%%%%%%%

\section{Introduction}
\label{sec:intro}
Consider a scenario where a search-and-rescue mission is underway, and rescue personnel needs to scan across hundreds of potential regions from a helicopter to locate a missing person. A crucial strategy in such operations involves using UAVs to capture aerial imagery that can help identify a target of interest (e.g.~the missing person). However, constraints like a limited field of view, high acquisition costs, time constraints, and restricted bandwidth between the sensor and the processing unit can make the search extremely challenging, demanding strategic decision making on where to query next based on the observations gathered so far. A similar challenge arises in other scenarios, such as locating a specific vehicle in an abduction case -- however, note that \emph{the target may differ, but the underlying problem structure remains the same}. In fact, many other scenarios share this general structure, such as anti-poaching enforcement \citep{fang2015security}, pinpointing landmarks, identifying drug or human trafficking sites, and more \citep{fang2016deploying,bondi2018airsim}. 

In this work, we derive and formalize a general task setup that encompasses these types of scenarios, and allows for controllable and reproducible model development and experimentation. We refer to our setup as \emph{Target-Conditioned Visual Active Search in Partially Observable environments (TC-POVAS)}, details of which are given in Sec.~\ref{sec:tc-povas}. The setup of TC-POVAS is as follows: Given a target category (or multiple target categories, depending on task requirements), the goal is to leverage a series of partially observed glimpses -- which are sequentially queried during active exploration -- to locate as many target objects as possible. Note that the number of allowed queries is assumed limited, to reflect factors such as time or resource constraints.

TC-POVAS builds on the visual active search (VAS) framework, where the aim is to find a target object using visual cues through sequential exploration~\citep{sarkar2023partially,sarkar2024geospatial}.
Past works assume access to a complete description of the search space (typically an aerial image that spans the whole area) for making decisions. However, in many real-world situations, e.g.~search-and-rescue operations, an entire view of the search space may not be available upfront. For example, an autonomous UAV on a rescue mission might only be able to capture partial glimpses through a series of narrow observations, due to confined viewing range and high data collection costs. In such cases, the agent has to make decisions with incomplete information, so models trained assuming access to complete images will struggle.

The challenge is twofold: \textbf{(i)} the agent must query the most informative patch from a partially observed scene to maximize information gain about the search space, and \textbf{(ii)} it must simultaneously ensure that this patch helps achieve the goal of locating the target objects. One might question why an agent cannot simply learn to choose patches that reveal target regions directly, without the need for acquiring knowledge about the underlying scene. The challenge arises because reasoning in unknown partially observable environments is inherently difficult. Thus, an agent must strike a balance between \emph{exploration} -- identifying patches that reveal the most information about the search space -- and \emph{exploitation} -- focusing on areas likely to contain target object(s) based on updated knowledge about the environment. An optimal agent must master this delicate balance to be effective. Additionally, previous VAS policies~\citep{sarkar2023partially,sarkar2024geospatial} are designed to search for specific target objects and cannot handle multiple categories simultaneously, which limits their adaptability to specific task preferences.

To address these challenges and to effectively tackle the TC-POVAS task setting, we propose \textbf{\emph{DiffVAS}}, a framework that consists of two key modules: \textbf{(1)} a diffusion-based \emph{conditional generative module (CGM)} and \textbf{(2)} a \emph{target-conditioned planning module (TCPM)}. The task of the CGM is to reconstruct an entire scene (search space) contingent on the partially observed glimpses gathered so far. To achieve this, we employ a neural network architecture that enables precise control over image generation by conditioning the diffusion-based generative model on the partially observed glimpses. Such a CGM attains fine control over image generation by integrating input conditions, like previously observed glimpses, directly into the model's intermediate layers, influencing the output at various stages of the diffusion process.
%This layered integration allows the model to align closely with the input conditions, ensuring that the generated image adheres to the desired structure while benefiting from the diffusion model's generative capabilities.

The objective of the TCPM is to decide which patch to query next by analyzing the partially observed glimpses together with the scene generated by the CGM, with the aim of revealing as many target regions as possible within the query budget. To accomplish this, the TCPM must learn to simultaneously explore the environment efficiently to maximize information gathering (exploration) \emph{and} select patches that reveal as many target regions as possible based on its current knowledge of the environment (exploitation). To this end, we develop an RL-based policy that learns to balance exploration and exploitation. To train the policy, we design a reward function that, besides encouraging target discovery, takes into account two key factors: \emph{local uncertainty} and \emph{global reconstruction quality}. These factors measure how effectively the policy issues actions that contribute to gaining information about the environment. 
Furthermore, we design the TCPM to be target-conditioned, which enables it to search for different target categories according to task requirements and handle multiple categories simultaneously. This is done by introducing an inference strategy that leverages target-conditioned probability distributions over grid cells for each target category, computed via TCPM, and learning target-aware state representation by leveraging cross-attention. Finally, we conduct extensive experiments to demonstrate the effectiveness of \emph{DiffVAS}.
\\ \\
\textbf{In summary, we make the following contributions:}
\begin{itemize}[noitemsep,topsep=0pt, leftmargin=*]
    \item We introduce TC-POVAS, a novel task setup that addresses target-conditional (TC) visual active search (VAS) in partially observable (PO) environments, and which extends traditional VAS to become more closely aligned with practical scenarios.
    \item We propose \emph{DiffVAS}, an agent that effectively tackles the TC-POVAS task by reconstructing the whole search area as it explores and searches for targets. Unlike previous approaches, DiffVAS can search a diverse range of target objects and tackle multiple target categories simultaneously.
    \item We demonstrate the significance of each component within DiffVAS through a comprehensive series of quantitative and qualitative ablation analyses.
    \item Extensive experimental evaluations using two publicly available satellite image datasets (xView, DOTA), across various unknown target settings, demonstrate that DiffVAS significantly outperforms all baseline approaches. Code and models will be made publicly available.
\end{itemize}
%\vspace{-4mm}

\section{TC-POVAS Task Setup}\label{sec:tc-povas}
\begin{figure}[t]
\includegraphics[width=0.95\linewidth]{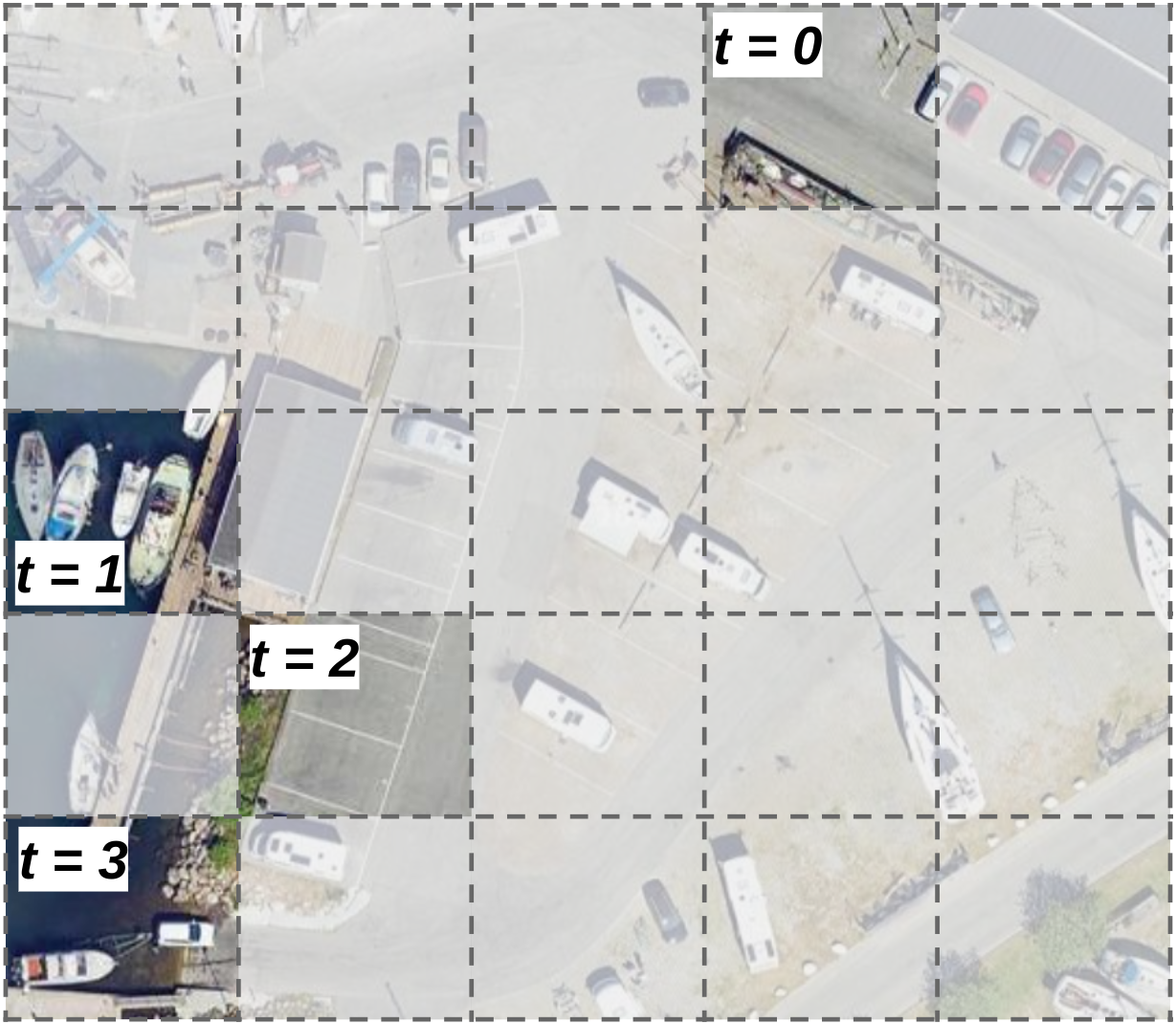}%,height=4cm
  \caption{The goal of TC-POVAS is to cover as many regions containing target instances as possible within a limited budget. In this case, the agent begins at the top of the search area at $t=0$ with a task budget $\mathcal{B}=3$, and is tasked with discovering boats and cars ($\mathcal{Z}=\{\text{boat}, \text{car}\}$). Its first action $a_1$ at $t=1$ leads it to discover a few boats ($y^{(a_1)}(\cdot \mid \mathcal{Z})=1$), at $t=2$ it does not discover any instances ($y^{(a_2)}(\cdot \mid \mathcal{Z})=0$), and at $t=3$ it discovers both a boat and a car ($y^{(a_3)}(\cdot \mid \mathcal{Z})=2$). The task now ends with an instance coverage score of $1+0+2=3$, as the budget has run out.}
  \label{fig:arch}
  %\vspace{-4mm}
\end{figure}

%\vspace{-2mm}
In this section, we describe the details of our proposed TC-POVAS task setup; see Fig.~\ref{fig:arch} for an overview. TC-POVAS is a search task in which one or multiple targets should be localized within a search area -- represented here as an aerial image $x$ that is partitioned into $N$ grid cells, such that $x = (x^{(1)}, x^{(2)}, ..., x^{(N)})$ -- within a given query budget $\mathcal{B}$, which here represents the number of movement actions. Each grid cell corresponds to a sub-image
%The search area is discretized into an X × Y grid superimposed on a given aerial landscape (i.e., $x$), with each grid corresponding to a position (location) 
and represents the limited field of view of the agent (akin to a UAV hovering at a limited altitude), i.e.~the agent can only observe the aerial content of a sub-image $x^{(i)}$ corresponding to the $i$th grid cell in which it is located at time step $t$. The agent's action space corresponds to all possible movements to other grid cells.
\begin{comment}
\end{comment}
For each task configuration, the target object categories are predefined in natural language, such as \enquote{small car, boat}, and represented as a set $\mathcal{Z}$. The objective is to uncover as many grid cells as possible that contain objects in $\mathcal{Z}$ by strategically exploring the grid cells within the budget constraint $\mathcal{B}$. To keep track of which grid cells $x^{(j)}$ contain targets, we label each grid cell $x^{(j)}$ with $y^{(j)}(\cdot \mid \mathcal{Z}) \in \{0, k\}$, where $y^{(j)}(\cdot \mid \mathcal{Z})=k$ if cell $j$ contains at least one instance each of $k$ different target object categories from set $\mathcal{Z}$, and 0 otherwise. The full label vector for the task is $y(\cdot \mid \mathcal{Z}) = (y^{(1)}(\cdot \mid \mathcal{Z}), y^{(2)}(\cdot \mid \mathcal{Z}), ..., y^{(N)}(\cdot \mid \mathcal{Z}))$. Naturally, at decision time we assume no direct knowledge of $y(\cdot \mid \mathcal{Z})$, but it is used to evaluate an agent's task performance at the end of an episode. Moreover, when an agent queries a grid cell $j$, it receives $x^{(j)}$ (the aerial image content of the $j$:th grid cell) and the ground truth label $y^{(j)}(\cdot \mid \mathcal{Z})$ for that cell.\footnote{It would also be possible to consider a setting where an aerial object detector is used to assess what objects are within a grid cell.}
See Appendix 5.11 for more on the task setup. Denoting a query performed in step $t$ as $q_t$ and $c(i, j)$ as the cost associated with querying grid cell $j$ starting from grid cell $i$, the task optimization objective is:
\begin{equation}\label{eq:agent-obj}
    %U(x^{\{(q_t)\}};\{q_t\}) \equiv 
    \max_{\{q_t\}} \sum_{t} y^{(q_t)}(\cdot \mid \mathcal{Z}) \quad\\
    \text{ subject to } \sum_{t\ge 0} c(q_{t-1},q_t) \le \mathcal{B}
\end{equation}
%For example, in a setup with constant movement cost $c(i,j)=1$, a budget of $\mathcal{C}=3$ and four grid cells containing 3, 2, 2, 1 different target categories, respectively, the solution to (\ref{eq:agent-obj}) is to query the grid cells containing 3, 2, 2 target categories (thus omitting the one with only one category, as not all target-containing queries can be done within budget).
\textbf{Target-Conditioned Partially Observable Markov Decision Process (TC-POMDP).}
With objective (\ref{eq:agent-obj}) in mind, we aim to learn a search policy that can efficiently explore a search area and discover target regions, and to achieve this through learning from similar pre-labeled search tasks, referred to as $\mathcal{D} = \{(x_i,y_i(\cdot \mid \mathcal{Z}))\}$, which consists of images $x_i$ paired with corresponding grid cell labels  $y_i(\cdot \mid \mathcal{Z})$. Here, each $x_i$ is composed of $N$ elements $(x_i^{(1)},x_i^{(2)},\ldots,x_i^{(N)})$ which represent the grid cells in the image, and each $y_i(\cdot \mid \mathcal{Z})$ contains $N$ corresponding labels $y_i^{(1)}(\cdot \mid \mathcal{Z}),y_i^{(2)}(\cdot \mid \mathcal{Z}),\ldots,y_i^{(N)}(\cdot \mid \mathcal{Z})$. We model this problem as a TC-POMDP and consider a family of TC-POMDP environments $\mathcal{M}^{e} = \{ (\mathcal{S}^{e}, \mathcal{A},\mathcal{X}^{e},\mathcal{T}^{e},\mathcal{G}^{e}, \gamma) | e \in \epsilon \}$, where $e$ is the environment index. Each environment $\mathcal{M}^{e}$ comprises a state space $\mathcal{S}^{e}$, shared action space $\mathcal{A}$, observation space $\mathcal{X}^{e} \in \{(x_e^{(1)},x_e^{(2)},\ldots,x_e^{(N)})\}$, transition dynamics $\mathcal{T}^{e}$, target space $\mathcal{G}^{e}(\mathcal{Z}) \subset \mathcal{S}^{e}$ such that $\mathcal{G}^{e}(\mathcal{Z}) = \{x_e^{(g)} \in \mathcal{X}^{e} \mid y_e^{(g)}(\cdot \mid \mathcal{Z}) \neq 0\ \text{for } g \in \{1, 2, \dots, N\}\}$, and discount factor $\gamma \in [0,1]$. $\mathcal{T}^{e}$ involve updating the remaining budget $\mathcal{B}_{t+1}$ by subtracting the current query cost $c(q_{t-1},q_t)$ and incorporating the latest query outcomes, i.e.~$x^{(q_t)}_e, y^{(q_t)}_e(\cdot \mid \mathcal{Z})$, into the state at time $t+1$. The observation $x^e \in \mathcal{X}^{e}$ is determined by state $s^e \in \mathcal{S}^{e}$ and the unknown environmental factor $b^e \in \mathcal{F}^e$, i.e.~$x^e(s^e, b^e)$, where $\mathcal{F}^e$ encompasses variations (including seasonality, weather effects, etc) related to diverse geospatial regions.
\begin{figure*}%[t]
%\vspace{-1mm}
  \includegraphics[width=1.0\linewidth]{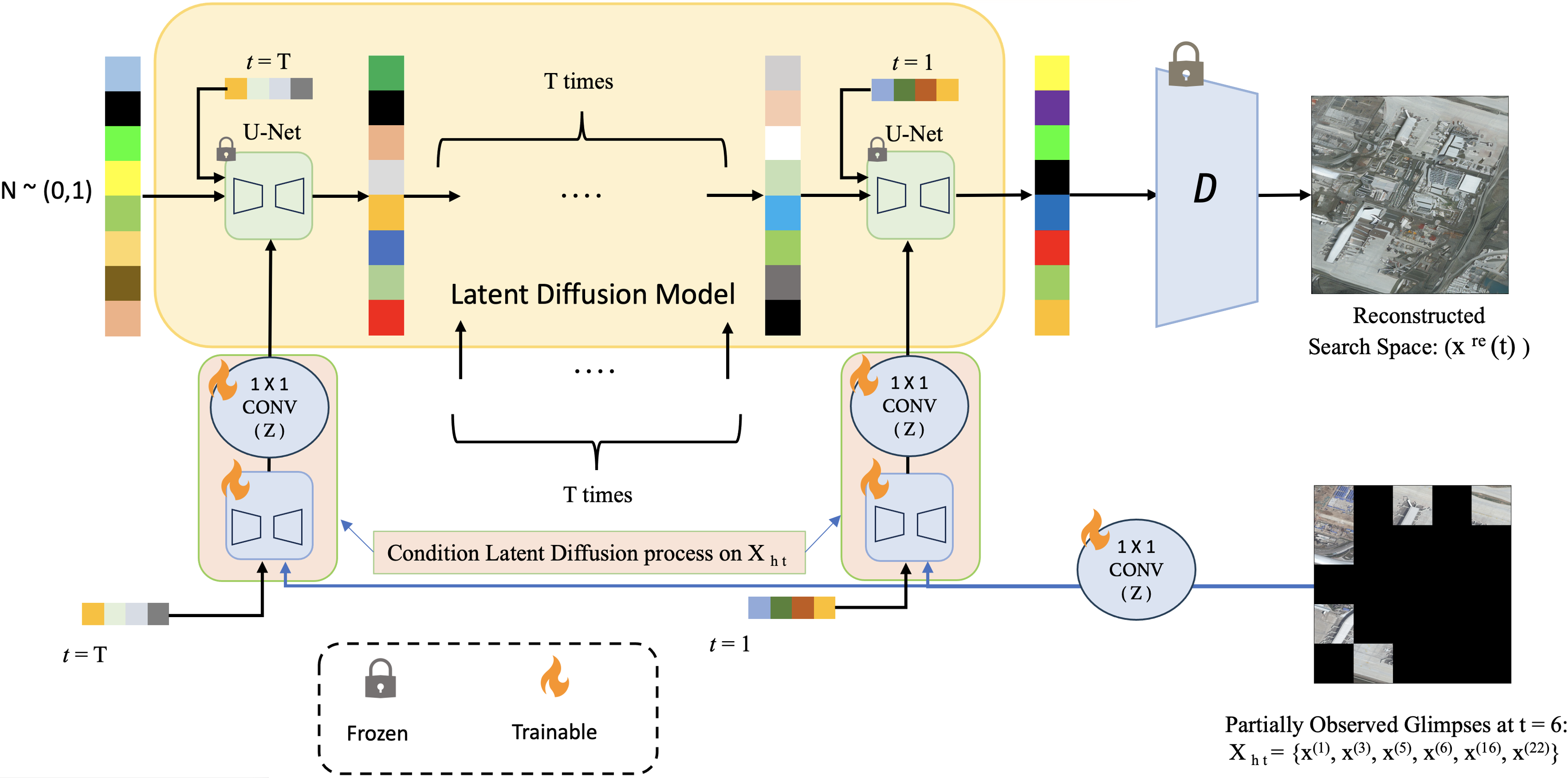}
  %\vspace{-3mm}
  %\caption{Conditional generative model architecture.}\label{fig:cgm}
  \caption{Overview of the conditional generative module (CGM) within DiffVAS. The diffusion-based CGM learns to reconstruct an entire search area based on a partially observed scene, which in turn helps guide subsequent decisions (by feeding the CGM's latent representation $l_{\text{re}}(t)$; see Fig.~\ref{fig:framework}) in order to maximize target discovery.}\label{fig:cgm}
  \label{fig:cgm_pretrain}
  %\vspace{-4mm}
\end{figure*}
Finally, $x^{(q_t)}_e$ denotes the observation associated with $q_t$ at step $t$, for domain $e$.

The primary objective in a TC-POMDP is to learn a history-aware target-conditioned policy $\pi(a_t | x^e_{h_t}, \mathcal{Z}, \mathcal{B}^{e}_t )$, where $x^e_{h_t} = (x^{(q_1)}_e, \ldots, x^{(q_{t})}_e)$ combines all the previous observations up to time $t$ and $\mathcal{B}^{e}_t$ represents the remaining budget at time, that maximizes the discounted state density function $J(\pi)$ across all domains $e \in \epsilon$, as follows:
\begin{comment}
\begin{equation}
\vspace{-3pt}
    J(\pi) = \mathbb{E}_{e \> \sim \> \epsilon,   \mathcal{B}^e_0 \> \sim \> \mathcal{B}^e, \mathcal{Z} \> \sim \> \text{RandomSubset}(\mathcal{O}^e), \pi} \left[ (1 - \gamma) \sum_{t=0}^{\infty} \gamma^t p_{\pi}^e (s_t \in \mathcal{G}^{e}(\mathcal{Z}) | \> \mathcal{Z}, \mathcal{B}^e_t ) \right]
\label{eq:rl_obj}
\end{equation}
\end{comment}
%\small
\begin{align}
    J(\pi) &= \mathbb{E}_{e \sim \epsilon, \mathcal{B}^e_0 \sim \mathcal{B}^e, 
    \mathcal{Z} \sim \text{RandomSubset}(\mathcal{O}^e), \pi} \Bigg[ (1 - \gamma) \nonumber \\
    &\quad \cdot \sum_{t=0}^{\infty} \gamma^t p_{\pi}^e (s_t \in \mathcal{G}^{e}(\mathcal{Z}) 
    \mid \mathcal{Z}, \mathcal{B}^e_t ) \Bigg]
\label{eq:rl_obj}
\end{align}

Here $p_{\pi}^e (s_t \in \mathcal{G}^{e}(\mathcal{Z}) | \> \mathcal{Z},  \mathcal{B}^e_t)$ represents the probability of querying a grid cell containing at least one target at step $t$ within domain $e$ under the policy $\pi( . | x^e_{h_t}, \mathcal{Z}, \mathcal{B}^e_t)$, $\mathcal{O}^e$ denotes the set of object categories in domain $e$, and $e \sim \epsilon, \mathcal{B}^e_0 \> \sim \> \mathcal{B}^e$ refer to uniform samples from each set. The total query budget allocated for a search task is denoted as $\mathcal{B}^e$. Throughout the training process, the agent is exposed to a set of training environments $\{e_i\}^N_{i=1} = \epsilon_{\text{train}} \subset \epsilon$, each identified by its environment index. To reduce clutter, we omit the notation $e$ for the rest of the paper. Next, we explore how we design and train a policy -- which we call \emph{DiffVAS} -- to effectively maximize the objective outlined in (\ref{eq:rl_obj}).

\section{DiffVAS: A Diffusion-Guided Approach for Tackling TC-POVAS}
In this section we introduce \emph{DiffVAS}, a diffusion-guided, reinforcement learning (RL)-based agent designed to address visual active search (VAS) in partially observable environments. DiffVAS is composed of two main modules: \textbf{(1)} a conditional generative module (CGM) and \textbf{(2)} a target-conditioned planning module (TCPM). Next, we detail each component of the proposed DiffVAS framework, starting with the training strategy for both modules to learn an efficient policy, followed by the inference procedure.
\subsection{Training}
%\vspace{-4pt}
Our approach uses a two-phase training strategy: In the first phase, we train the $\mathrm{CGM}$, and then we freeze its parameters while training the $\mathrm{TCPM}$ in the second phase. The purpose of the $\mathrm{CGM}$ is to synthesize the entire scene (i.e.~the search space) from the partially observed glimpses collected so far, thereby assisting the $\mathrm{TCPM}$ in deciding the next query location. To achieve this, the conditional generative model leverages a diffusion-based adapter-style approach~\citep{mou2024t2i,zhang2023adding}. Diffusion models are powerful generative models that allow for precise control over the attributes of the generated samples. While these diffusion models trained on large datasets have achieved success, there is often a need to introduce additional controls in downstream fine-tuning processes. In our case, the $\mathrm{CGM}$ fine-tunes the diffusion model by integrating information about previously observed glimpses $x_{h_t}$ while preserving the integrity of the pre-trained diffusion model. This is done by freezing the parameters of a trained diffusion model and creating a trainable copy that takes an external conditioning vector $x_{h_t}$ as input (see Fig.~\ref{fig:cgm_pretrain}). The trainable copy is connected to the frozen pre-trained diffusion model using zero convolution layers $Z(;)$, which are $1 \times 1$ convolution layers initialized with weights and biases set to zero, which safeguards the model against any harmful noise in the early stages of training, as outlined in~\citep{zhang2023adding}. 
This design strategy thus retains the capabilities of the large-scale pre-trained diffusion model while allowing the trainable copy to adapt to new conditions.
\\ \\
\textbf{CGM training.} To train the parameters of the $\mathrm{CGM}$, we randomly sample an image $x_0$ corresponding to an entire search space, and progressively add noise to create a noisy image $x_k$, where $k$ indicates the number of noise additions.

\begin{figure*}[t]
    \centering
    \includegraphics[width=0.93\linewidth]{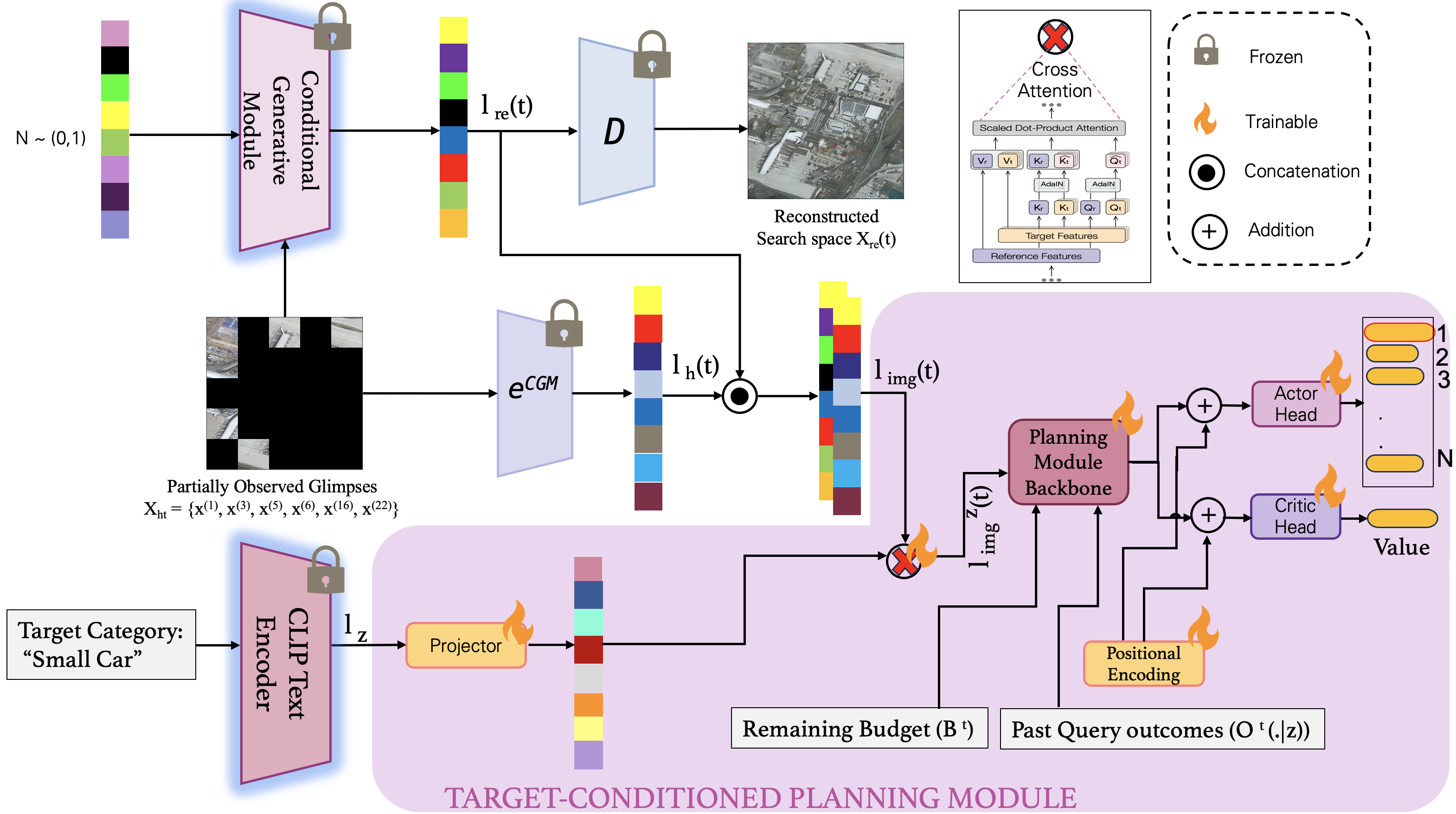}
    %\vspace{-4mm}
    \caption{The proposed DiffVAS framework for diffusion-guided visual active search in partially observable environments.}
    %\vspace{-5mm}
    \label{fig:framework}
\end{figure*}

Conditioned on partially observed glimpses $x_{h_t}$, CGM trains a network $\epsilon_{\theta}$ to predict the noise added to $x_k$ as follows:
\begin{equation}
    \mathcal{L}_{CGM} = \mathbb{E}_{x_0, k, x_{h_t}, \epsilon \sim \mathcal{N}(0,1)} \left[ \|\epsilon - \epsilon_\theta(x_k, k, x_{h_t})\|_2^2 \right] 
\end{equation}
%$\mathcal{L}_{CGM}$ represents the overall learning objective of the $\mathrm{CGM}$.
Note that $x_{h_t}$ is obtained by randomly selecting a history length $h_t \in \{1, \ldots, N-1\}$, then choosing $h_t$ random patches while masking the rest of $x_0$. An overview of the $\mathrm{CGM}$ is shown in Fig.~\ref{fig:cgm_pretrain}.
%; see architecture and hyperparameter details in the appendix.
%We next discuss the training procedure for the $\mathrm{TCPM}$.
%%%%%%%%%%%%%%%%%%%%%%%%%%%%%%%%%%%%%%%%%%%%%%%%%%%%%%%%%%%%
\\ \\
\textbf{TCPM training.} The role of the $\mathrm{TCPM}$ is to determine the next query location based on $x_{h_t}$, $\mathcal{B}^t$, and the target category $\mathcal{Z}$. The planning module must \emph{explore} -- seeking patches that provide the most insight into the search space -- while also \emph{exploiting} known information, focusing on areas with a high likelihood of containing the target. To this end, we develop an actor-critic style PPO algorithm~\citep{schulman2017proximal} for learning a policy that balances exploration and exploitation. Since decision-making in an unknown environment is challenging, we leverage the trained CGM to reconstruct the entire search space $x_{\text{re}}(t)$ from partially observed glimpses $x_{h_t}$. This reconstructed information aids the planning module in making more informed decisions about the next query location.  As illustrated in Fig.~\ref{fig:framework}, the latent representation $l_{\text{re}}(t)$ of $x_{\text{re}}(t)$ is extracted from the encoder at the final step of the reverse diffusion process of the pre-trained CGM (i.e., $x_{\text{re}}(t) = D(l_{\text{re}}(t) = \mathrm{CGM}(x_{h_t}))$.
%is extracted at the very last step of the reverse diffusion process through the encoder of the pre-trained conditional generative module ($E_{CGM}$), forming a key component of the input state for the planning module.
We use the encoder $e^{\mathrm{CGM}}(\cdot)$ of the CGM as a feature extractor to derive the latent representation $l_{h}(t)$ of $x_{h_t}$, i.e.~$l_{h}(t) = e^{\mathrm{CGM}}(x_{h_t})$. We merge $l_{\text{re}}(t)$ and $l_{h}(t)$ channel-wise, forming the combined representation $l_{\mathrm{img}}(t)$. The key reason for incorporating $l_{h}(t)$ into the state space is that early in the search, the reconstruction $x_{\text{re}}(t)$ of the search space may be unreliable, making it imprudent to base decisions solely on $l_{\mathrm{re}}(t)$.

As we want to learn a policy capable of searching for diverse target objects, we condition it on the target object $z$. Here, $z$ is an element of the set of target object categories (i.e.~$z \in \mathcal{Z}$; see Sec.~\ref{sec:inf} for how the multi-target setting is handled). The target object embedding $l_{z}$ is obtained via the CLIP~\citep{radford2021learning} text encoder (i.e., $l_{z} = f^{\mathrm{CLIP}}(z)$). A learnable cross-attention layer is then applied between $l_{z}$ and $l_{\mathrm{img}}(t)$, which yields a representation $l^{z}_{\mathrm{img}}(t)$ of the search space that is target-aware. At time $t$, the planning module's input state comprises $l_{\mathrm{img}}(t)$, $l_{z}$, the remaining budget $\mathcal{B}^{t}$, and an observation vector $o^t(\cdot \mid z)$ that encodes previous search query outcomes. Each element of $o^t(\cdot \mid z)$ corresponds to a grid cell index, where $o^t_{(j)}(\cdot \mid z) = 2 y^{(j)}(\cdot \mid z) - 1$ if the $j$:th grid cell has been explored, and $o^t_{(j)}(\cdot \mid z) = 0$ otherwise. The primary reason for incorporating $\mathcal{B}^{t}$ and $o^t(\cdot \mid z)$ into the state space is to ensure that the planning module makes decisions with full awareness of both remaining budget and previous query outcomes. 

Denote the state at time $t$ as $s_t = [l_{\mathrm{img}}(t), l_{z}, o^t(\cdot \mid z), \mathcal{B}^{t}]$. Training $\mathrm{TCPM}$ is done using PPO \cite{schulman2017proximal} and involves learning both an \emph{actor} (policy network, parameterized by $\zeta$) $\pi_\zeta: s_t \xrightarrow{} p(\mathcal{A})$ and a \emph{critic} (value network, parameterized by $\eta$) $\mathit{V}_\eta: s_t \xrightarrow{} \mathbb{R}$ that approximates the true value $V^{\mathrm{true}} ( s_t) = \mathbb{E}_{a \sim \pi_\zeta (.|l_{\mathrm{img}}(t), l_{z}, o^t(\cdot \mid z), \mathcal{B}^{t})} [R(s_t, a_t, z) + \gamma V(\mathcal{T} (s_t, a_t))]$. We optimize both the actor and critic networks with the following loss:
\begin{align}
\mathcal{L}^{\mathrm{planner}}_t(\zeta, \eta) &= 
    \mathbb{E}_t \Big[ 
    -\mathcal{L}^{\mathrm{clip}}(\zeta) 
    + \alpha \mathcal{L}^{\mathrm{crit}}(\eta) \nonumber \\
    &\quad - \beta \mathcal{H} \big[ \pi_\zeta(\cdot \mid l_{\mathrm{img}}(t), l_{z}, 
    o^t(\cdot \mid z), \mathcal{B}^{t}) \big] 
    \Big]
    \label{eq:ppo}
\end{align}
Here $\alpha$ and $\beta$ are hyperparameters, and $\mathcal{H}$ denotes entropy, so minimizing the final term of (\ref{eq:ppo}) encourages the actor to exhibit more exploratory behavior. The $\mathcal{L}^{\mathrm{crit}}$ loss is used to optimize the critic network and is defined as a squared-error loss, i.e.~$\mathcal{L}^{\mathrm{crit}}=(V_\eta(l_{\mathrm{img}}(t), l_{z}, o^t(\cdot \mid z), \mathcal{B}^{t}) - V^{\mathrm{true}} (s_t))^{2}$. The clipped surrogate objective $\mathcal{L}^{\mathrm{clip}}$ is employed to optimize the parameters of the actor-network while constraining the change to a small value $\epsilon$ relative to the old actor policy $\pi^{\mathrm{old}}$ and is defined as:
\begin{comment}
\begin{small}
\begin{align}
    \mathcal{L}^{\mathrm{clip}}(\zeta)=\mathrm{min}\left\lbrace\frac{\pi_\zeta(.| l_{\mathrm{img}}(t), l_{z}, o^t(\cdot \mid z), \mathcal{B}^{t})}{\pi^{\mathrm{old}}(.| l_{\mathrm{img}}(t), l_{z}, o^t(\cdot \mid z), \mathcal{B}^{t})} A^t, \mathrm{clip}\left(1-\epsilon, 1+\epsilon, \frac{\pi_\zeta(.| l_{\mathrm{img}}(t), l_{z}, o^t(\cdot \mid z), \mathcal{B}^{t})}{\pi^\mathrm{old}(.| l_{\mathrm{img}}(t), l_{z}, o^t(\cdot \mid z, \mathcal{B}^{t})} ) A^t\right)\right\rbrace\nonumber\\
    A^t = r_t + \gamma r_{t+1} + \ldots + \gamma^{T-t+1} r_{T-1}  - V_\eta (l_{\mathrm{img}}(t), l_{z}, o^t(\cdot \mid z), \mathcal{B}^{t})
\end{align}
\end{small}
\end{comment}
%\small
\begin{align}
    \mathcal{L}^{\mathrm{clip}}(\zeta) &= \min \Bigg\lbrace 
    \frac{\pi_\zeta(\cdot \mid l_{\mathrm{img}}(t), l_{z}, o^t(\cdot \mid z), \mathcal{B}^{t})}
    {\pi^{\mathrm{old}}(\cdot \mid l_{\mathrm{img}}(t), l_{z}, o^t(\cdot \mid z), \mathcal{B}^{t})} A^t, \nonumber \\
    &\quad \text{clip} \Big(1-\epsilon, 1+\epsilon, 
    \frac{\pi_\zeta(\cdot \mid l_{\mathrm{img}}(t), l_{z}, o^t(\cdot \mid z), \mathcal{B}^{t})}
    {\pi^\mathrm{old}(\cdot \mid l_{\mathrm{img}}(t), l_{z}, o^t(\cdot \mid z), \mathcal{B}^{t})} 
    \Big) A^t \Bigg\rbrace \nonumber \\
    A^t &= r_t + \gamma r_{t+1} + \dots + \gamma^{T-t+1} r_{T-1}  \nonumber \\
    &\quad - V_\eta (l_{\mathrm{img}}(t), l_{z}, o^t(\cdot \mid z), \mathcal{B}^{t})
\end{align}
After every fixed update step, we copy the parameters of the current policy network $\pi_\zeta$ onto the old policy network $\pi^{\mathrm{old}}$ to enhance training stability.
%All hyperparameter details for training the actor and critic network are in the appendix.
Our proposed DiffVAS framework is illustrated in Fig.~\ref{fig:framework}. Next, we introduce a novel reward function $\mathcal{R}$ designed to guide the planning module in mastering search strategy in partially observed scenes.
\\ \\
\textbf{Reward structure.} The reward $\mathcal{R}$ consists of three components: \textbf{(i)} \emph{local uncertainty} reward $\mathcal{R}^{\mathrm{LU}}$, \textbf{(ii)} \emph{global reconstruction} reward $\mathcal{R}^{\mathrm{GR}}$, and \textbf{(iii)} \emph{active search} reward $\mathcal{R}^{\mathrm{AS}}$. The $\mathcal{R}^{\mathrm{LU}}$ and $\mathcal{R}^{\mathrm{GR}}$ rewards assess how efficiently the planning module's movements enhance information-gathering about the environment (\emph{exploration}), and $\mathcal{R}^{\mathrm{AS}}$ assesses how well the policy is discovering target regions (\emph{exploitation}). We define $\mathcal{R}^{\mathrm{LU}}$ as:
\begin{comment}
{\small
\begin{equation*}
\mathcal{R}^{\mathrm{LU}}=\mathrm{sgn}\left[\left\{\mathrm{SSIM}\left(x_{\mathrm{true}}^{(a_{\mathrm{ran}})},D\left(\mathrm{CGM}(x_{h_{t-1}})\right)^{(a_{\mathrm{ran}})}\right)\right\} - \left\{\mathrm{SSIM}\left( x_{\mathrm{true}}^{(a_t)},D\left(\mathrm{CGM}(x_{h_{t-1}})\right)^{(a_{t})}\right)\right\} \right]
\label{eq:local_unc}
\end{equation*}}
\end{comment}
%\small
\begin{align}
\mathcal{R}^{\mathrm{LU}} = \mathrm{sgn} \Big[ & \mathrm{SSIM}\big(x_{\mathrm{true}}^{(a_{\mathrm{ran}})}, 
D(\mathrm{CGM}(x_{h_{t-1}}))^{(a_{\mathrm{ran}})}\big)  \nonumber \\
& - \mathrm{SSIM} \big(x_{\mathrm{true}}^{(a_t)}, 
D(\mathrm{CGM}(x_{h_{t-1}}))^{(a_t)}\big) \Big]
\label{eq:local_unc}
\end{align}
where the structural similarity index~\citep{wang2002universal} $\mathrm{SSIM}(a, b)$ is used to measure the similarity between two images $a$ and $b$; $a_{\mathrm{ran}}$ represents a randomly selected grid cell at time $t$; {\small $ x_{\mathrm{true}}^{(a_{\mathrm{ran}})}$} and {\small $x_{\mathrm{true}}^{(a_t)}$} refer to the $a_{\mathrm{ran}}$:th and $a_t$:th grid cells of the ground truth image, respectively. According to (\ref{eq:local_unc}), the agent receives a positive reward when the ground truth and reconstructed patches are more \emph{dissimilar} (according to $\mathrm{SSIM}$) for the queried grid cell than for a randomly selected grid cell (i.e., $a_{\mathrm{ran}}$). Thus, (\ref{eq:local_unc}) gives a positive reward when the agent queries a patch that it is uncertain of, which encourages the discovery of novel (and uncertain) parts of the search area.
The global reconstruction reward is defined as:
\begin{comment}
\begin{equation}
\mathcal{R}^{\mathrm{GR}} = \mathrm{sgn}\left[\left\{\mathrm{SSIM}\left(x_{\mathrm{true}},D\left(\mathrm{CGM}(x_{h_t})\right)\right)\right\} -  \left\{\mathrm{SSIM}\left(x_{\mathrm{true}},D\left(\mathrm{CGM}(x_{h_t^{\mathrm{ran}}})\right)\right)\right\} \right]
\label{eq:global_re}
\end{equation}
\end{comment}
%\small
\begin{align}
\mathcal{R}^{\mathrm{GR}} = \mathrm{sgn} \Big[ & \mathrm{SSIM} \big( x_{\mathrm{true}}, 
D(\mathrm{CGM}(x_{h_t})) \big) \nonumber \\
& - \mathrm{SSIM} \big( x_{\mathrm{true}}, 
D(\mathrm{CGM}(x_{h_t^{\mathrm{ran}}})) \big) \Big]
\label{eq:global_re}
\end{align}
where $x_{h_t^{\mathrm{ran}}}$ is identical to $x_{h_t}$, except the action $a_t$ at time $t$ is replaced with a random action $a_{\mathrm{ran}}$. As seen in (\ref{eq:global_re}), $\mathcal{R}^{\mathrm{GR}}$ rewards the agent if querying the grid cell ($a_t$) results in a \emph{better} reconstruction of the entire search space by the CGM module compared to querying a random grid cell ($a_{\mathrm{ran}}$) -- thus note that this reward term is in some sense \enquote{inverse} relative to (\ref{eq:local_unc}). In the early stages of the search, the search space reconstruction by CGM is poor (see an example in Fig.~\ref{fig:vis_tc-povas}) regardless of the queried grid cell, making the $\mathcal{R}^{\mathrm{GR}}$ reward signal weak. Therefore, relying solely on $\mathcal{R}^{\mathrm{GR}}$ is not effective for distinguishing between good and bad grid cell selections. In this scenario, $\mathcal{R}^{\mathrm{LU}}$ offers a sharper distinction, as it is based on evaluating a single grid cell.

To ensure the agent's queried grid cell also contributes to identifying regions containing target objects $z$, we design an active search reward function $\mathcal{R}^{\mathrm{AS}}$ defined as $\mathcal{R}^{\mathrm{AS}} = y^{(a_t)}(\cdot \mid z)$ if the agent visits an unexplored cell and $\mathcal{R}^{\mathrm{AS}} = -5$ otherwise (penalizing the agent heavily for querying a grid cell more than once). Thus, if the agent moves to an unexplored cell it receives a reward $+1$ if it contains a target, and $0$ otherwise.
Finally, the total reward used when training is given by:
\begin{equation}
    \mathcal{R}(s_t, a_t, z) = \mathcal{R}^{\mathrm{LU}} + \mathcal{R}^{\mathrm{GR}} + \mathcal{R}^{\mathrm{AS}}
    \label{eq:reward}
\end{equation}
%Next, we discuss the inference procedure of DiffVAS.

\subsection{Inference}\label{sec:inf}
%Since each task has a distinct objective, it requires searching for different sets of target objects during inference.
In this section we explain how a trained DiffVAS agent is used to search for one or multiple target categories simultaneously, based on task requirements. Denote the set of target object categories to be searched as $\mathcal{Z} = \{ z_1, \ldots, z_k\}$ (e.g.~'car' ($z_1$), 'boat' ($z_2$), and so on). At each step, we first compute $k$ individual probabilities of querying each grid cell, conditioned on the $c$:th category $z_c \in \mathcal{Z}$, i.e.~$p_c=\pi_\zeta(.| l_{\mathrm{img}}(t), l_{z_c}, o^t(\cdot \mid z_c), \mathcal{B}^{t})$ for each $c \in \{1,\dots,k\}$. We then select the next grid cell to query based on the joint probability distribution, defined as:
%\vspace{-2mm}
%\small
\begin{align}
    \pi_\zeta\left(\cdot \mid l_{\mathrm{img}}(t), l_{\mathcal{Z}}, o^t(\cdot \mid \mathcal{Z}), \mathcal{B}^{t}\right) 
    &= \prod_{c=1}^{k} p_c%, \nonumber %\\
    %\text{where} \quad p_i = \pi_\zeta(\cdot \mid l_{\mathrm{img}}(t), &l_{z_i}, o^t(\cdot \mid z_i), \mathcal{B}^{t})
    \label{eq:inference}
\end{align}
%\vspace{-mm}
%Here, $\mathcal{Z}$ denotes the set of target categories specified in natural language (e.g., $\mathcal{Z}$ =$ \{ \mathrm{car}, \mathrm{truck}, \mathrm{boat} \}$), while $z_\mathrm{c}$ represents an individual category within this set.
Thus, note that our proposed inference approach enables DiffVAS to flexibly handle tasks with varying numbers of target categories, overcoming a key limitation of previous VAS frameworks. We detail our inference process in Algorithm~\ref{A:algo}.

\begin{algorithm}
%\small
\caption{Inference procedure of \textsc{DiffVAS}}\label{alg:PARVS}
\begin{algorithmic}[1]  
\Require  Task instance with initial observation$(x^{\mathrm{(init)}}, y^{\mathrm{(init)}})$; set of target objects $\mathcal{Z} = \{ z_1, \ldots, z_k\}$; budget $\mathcal{B}$; trained CGM; encoder $e^{\mathrm{CGM}}$ of CGM; CLIP text encoder $f^{\mathrm{CLIP}}$; trained TCPM parameters ($\zeta$, $\eta$). 
\State \textbf{Initialize} $o^{0}(\cdot \mid z_{c}) = [ 0, \dots, 0] \>\text{for each} \>c \in \{1, \ldots, k\}; \mathcal{B}^{0} = \mathcal{B}$; $x_{h_0} = \{ x^{\mathrm{(init)}}\};  \ \text{step} \ t = 0;  R^{\mathrm{task}}=0$
\While{$\mathcal{B}^{t} > 0 $}
    
    %\State $\tilde{y} = f_{\theta_i}(x_i, o^{t})$
    \State $l_{\mathrm{img}}(t) = \mathrm{CGM}(x_{h_t}) \oplus e^{\mathrm{CGM}}(x_{h_t})$, where $\oplus$ represents channel-wise concatenation operation.
    \For{$c = 1$ to $k$}
        \State Compute $l_{z_c} = f^{\mathrm{CLIP}}(z_c)$,  and $p_c = \pi_\zeta(\cdot \mid l_{\mathrm{img}}(t), l_{z_c}, o^t(\cdot \mid z_{c}), \mathcal{B}^{t})$
    \EndFor
    \State Sample next grid cell index $j \sim \prod_{c=1}^{k} p_c$
    \State Query grid cell with index $j$ and observe $x^{(j)}$ and true label $y^{(j)} = \{y^{(j)}(\cdot \mid z_{1}), \ldots, y^{(j)}(\cdot \mid z_{k})\}$.
    \State Obtain $R^t = \sum_{c=1}^{k} y^{(j)}(\cdot \mid z_{c})
$; update $o^{t}_{(j)}(\cdot \mid z_{c})$ with $o^{t+1}_{(j)}(\cdot \mid z_{c}) = 2y^{(j)}(\cdot \mid z_{c}) -1$ (for each $c \in \{1, \ldots, k \}$), and update $\mathcal{B}^{t}$ with $\mathcal{B}^{t+1} = \mathcal{B}^{t} - c(k,j)$ (assuming we query the $k$:th grid at $(t-1)$).
    \State $R^{\mathrm{task}} = R^{\mathrm{task}} + R^{\mathrm{t}}$; incorporate latest observation $x^{(j)}$ into $x_{h_t}$, i.e.~$x_{h_{t+1}} = \{ x_{h_t}, x^{(j)} \}$. 
    \State \emph{t} $\xleftarrow{} t+1$
\EndWhile
\State \textbf{Return} $R^{\mathrm{task}}$
\end{algorithmic}
\label{A:algo}
\end{algorithm}
%\vspace{-15pt}

% (1) Time step of diffusion is not same as time step of planning; 

\vspace{-3mm}
\section{Experiments and Results}
\noindent
\textbf{Evaluation metrics.}
%\textbf{Evaluation metrics.}
Since VAS aims to maximize the identification of patches with target objects, we evaluate performance using the \emph{average number of targets (ANT)} identified through exploration in partially observable environments. 
%Although ANT evaluation metric alone is not fully indicative and can not fully capture the efficiency of a search policy.
%ANT alone is insufficient to fully capture the efficiency of a search policy as each target category appears with varying frequency in the search space. Hence, we also evaluate performance using Recall (denoted as REC), defined as 
In this work, we focus primarily on uniform query costs, i.e.~$c(i, j) = 1$ for all $i, j$, so $\mathcal{B}$ represents the total number of queries. Hence, ANT is defined as: 
\vspace{-1mm}
\begin{equation}
\mathrm{ANT} = \frac{1}{L}\sum_{i=1}^{L}\sum_{t=1}^{\mathcal{B}} y_{i}^{(q_t)}(\cdot \mid \mathcal{Z}) \text{ where } L = \text{number of test tasks }
\end{equation}
%\vspace{-1mm}
We evaluate DiffVAS and baselines for varying search budgets $\mathcal{B} \in \{ 5, 7, 10\}$ on a $5 \times 5$ grid structure.
%In the appendix, we conduct additional experiments for various grid
%configurations, each employing different values of $\mathcal{B}$ with varying target sets $\mathcal{Z}$.
\\ \\
\textbf{Baselines.} 
We compare our proposed DiffVAS policy to the following baselines: \textbf{(i)} Random Search (RS) selects unexplored grid cells at uniform random; \textbf{(ii)} E2EVAS~\citep{sarkar2024visual} is an RL-based approach for VAS in a fully observable space; \textbf{(iii)} Meta Partially Supervised VAS (MPS-VAS)~\citep{sarkar2023partially} is the state-of-the-art RL-based approach for single-target VAS, and is designed to learn an adaptable policy in a fully observable space.
\\ \\
\textbf{Datasets.} 
We evaluate DiffVAS and the baselines on two datasets: xView~\citep{lam2018xview} and DOTA~\citep{xia2018dota}. Both xView and DOTA are satellite image datasets, with roughly 3000 px per dimension and representing approximately 60 object categories. We use 50\%, 17\%, and 33\% of the large satellite images to train, validate, and test the methods, respectively. In the main paper, we compare the performance of DiffVAS with the baselines using the DOTA dataset.
%Similar results for xView are presented in the appendix.
\\ \\
\textbf{Single-category search tasks.}
%single target category evaluation
We begin by considering a setting with $\mathcal{Z}$ containing a single target category, as in most prior works. We evaluate the proposed methods with the following target classes: Large Vehicle (LV), Helicopter, Ship, Plane, Roundabout, and Harbor. The results are presented in Table~\ref{tab: dota_single_target}.
\begin{table}[b]
    \centering
    \small
    %\vspace{-2pt}
    \caption{ANT comparisons on the DOTA dataset for the single-target category setting. DiffVAS consistently performs best.}
    \vspace{-2mm}
    \setlength{\tabcolsep}{2.2pt} % Adjust column spacing for better fit
    \begin{tabular}{p{1.20cm}p{0.79cm}p{0.73cm}p{0.86cm}p{0.73cm}p{0.73cm}p{0.86cm}p{0.73cm}p{0.73cm}p{0.86cm}}
        \toprule
        \multicolumn{4}{c}{Test with $\mathcal{Z}$ = \{ Ship \}} & 
        \multicolumn{3}{c}{Test with $\mathcal{Z}$ = \{ LV \}} & 
        \multicolumn{3}{c}{Test with $\mathcal{Z}$ = \{ Plane \}} \\
        \midrule
        Method & $\mathcal{B}=5$ & $\mathcal{B}=7$ & $\mathcal{B}=10$ & 
        $\mathcal{B}=5$ & $\mathcal{B}=7$ & $\mathcal{B}=10$ & 
        $\mathcal{B}=5$ & $\mathcal{B}=7$ & $\mathcal{B}=10$ \\
        \cmidrule(r){1-4} \cmidrule(r){5-7} \cmidrule(l){8-10} 
        RS & 1.68 & 2.23 & 3.24 & 2.05 & 2.76 & 4.88 & 2.11 & 2.95 & 3.92 \\
        E2EVAS & 1.73 & 2.47 & 3.52 & 2.19 & 3.11 & 4.91 & 2.42 & 3.14 & 4.01 \\
        MPS-VAS & 1.77 & 2.50 & 3.59 & 2.22 & 3.15 & 4.96 & 2.53 & 3.17 & 4.08 \\
        \hline 
        \textbf{DiffVAS} & \textbf{2.12} & \textbf{3.22} & \textbf{3.91} & 
        \textbf{2.54} & \textbf{3.57} & \textbf{5.78} & 
        \textbf{3.12} & \textbf{4.07} & \textbf{5.24} \\
        \hline
        \hline
        \multicolumn{4}{c}{Test with $\mathcal{Z}$ = \{ Harbor \}} & 
        \multicolumn{3}{c}{Test with $\mathcal{Z}$ = \{ Roundabout \}} & 
        \multicolumn{3}{c}{Test with $\mathcal{Z}$ = \{ Helicopter \}} \\
        \midrule
        Method & $\mathcal{B}=5$ & $\mathcal{B}=7$ & $\mathcal{B}=10$ & 
        $\mathcal{B}=5$ & $\mathcal{B}=7$ & $\mathcal{B}=10$ & 
        $\mathcal{B}=5$ & $\mathcal{B}=7$ & $\mathcal{B}=10$ \\
        \cmidrule(r){1-4} \cmidrule(l){5-7} \cmidrule(l){8-10}
        RS & 1.56 & 2.43 & 3.67 & 1.54 & 2.83 & 4.04 & 1.32 & 3.15 & 4.56 \\
        E2EVAS & 1.68 & 2.57 & 3.90 & 1.77 & 2.97 & 4.18 & 1.61 & 3.29 & 4.61 \\
        MPS-VAS & 1.73 & 2.63 & 3.96 & 1.86 & 3.01 & 4.25 & 1.70 & 3.44 & 4.78 \\
        \hline
        \textbf{DiffVAS} & \textbf{2.01} & \textbf{3.15} & \textbf{4.45} & 
        \textbf{2.32} & \textbf{3.33} & \textbf{4.89} & 
        \textbf{2.12} & \textbf{3.91} & \textbf{5.05} \\
        \bottomrule
    \end{tabular}
    \label{tab: dota_single_target}
    %\vspace{-4mm}
\end{table}
We observe significant improvements in the performance of the proposed DiffVAS approach compared to all baselines in each different target setting, ranging from 8.9\% to 28.8\% improvement relative to the most competitive MPS-VAS method.

In each target setting, search performance improves as $\mathcal{B}$ increases, with DiffVAS typically gaining a greater advantage over other baselines. As more patches are revealed, the CGM-based reconstruction becomes more accurate, allowing DiffVAS to better exploit the search space and further enhance its search policy with a larger search budget $\mathcal{B}$. The importance of $\mathrm{TCPM}$ is demonstrated by the superior performance of  DiffVAS across all diverse target categories, as presented in Table~\ref{tab: dota_single_target}.
\\ \\
\textbf{Multi-category search tasks.} Next, we evaluate the proposed DiffVAS with $\mathcal{Z}$ encompassing multiple target categories and present the results in Table~\ref{tab: dota_multi_target}. We observe a substantial performance boost for DiffVAS across various target category sets, ranging from 8.3\% to 48.8\% improvement relative to the most competitive baseline, highlighting the effectiveness of our proposed inference strategy.
\begin{table}
    \centering
    \small
    \caption{ANT comparisons on the DOTA dataset for the multiple-target category setting. DiffVAS outperforms the other methods.}
    \vspace{-2mm}
    \setlength{\tabcolsep}{2.2pt} % Adjust column spacing for better fit
    \begin{tabular}{p{1.20cm}p{0.73cm}p{0.73cm}p{0.86cm}p{0.76cm}p{0.73cm}p{0.86cm}p{0.73cm}p{0.73cm}p{0.86cm}}
        \toprule
        \multicolumn{4}{c}{Test with $\mathcal{Z}$ = \{ Ship, Harbor \}} & 
        \multicolumn{3}{c}{Test with $\mathcal{Z}$ = \{ LV, SV \}} & 
        \multicolumn{3}{c}{Test with $\mathcal{Z}$ = \{ Plane, Helicopter\}} \\
        \midrule
        Method & $\mathcal{B}=5$ & $\mathcal{B}=7$ & $\mathcal{B}=10$ & 
        $\mathcal{B}=5$ & $\mathcal{B}=7$ & $\mathcal{B}=10$ & 
        $\mathcal{B}=5$ & $\mathcal{B}=7$ & $\mathcal{B}=10$ \\
        \cmidrule(r){1-4} \cmidrule(r){5-7} \cmidrule(l){8-10}
        RS & 2.34 & 3.19 & 4.12 & 2.31 & 3.67 & 4.91 & 1.99 & 3.90 & 5.26 \\
        E2EVAS & 2.37 & 3.22 & 4.14 & 2.33 & 3.71 & 4.93 & 2.04 & 3.95 & 5.30 \\
        MPS-VAS & 2.38 & 3.26 & 4.18 & 2.38 & 3.72 & 4.97 & 2.09 & 3.98 & 5.33 \\
        \hline
        \textbf{DiffVAS} & \textbf{2.98} & \textbf{4.16} & \textbf{4.92} & 
        \textbf{3.05} & \textbf{4.33} & \textbf{5.52} & 
        \textbf{3.11} & \textbf{4.34} & \textbf{6.02} \\
        \bottomrule
    \end{tabular}
    \label{tab: dota_multi_target}
    \vspace{-4mm}
\end{table}
Note that, as shown in Tables~\ref{tab: dota_single_target} and~\ref{tab: dota_multi_target}, ANT values vary across different $\mathcal{Z}$ because each target category appears with different frequencies in the search space.
%Next, we analyze each module within DiffVAS.
\\ \\
\textbf{Importance of the CGM.} To investigate the significance of the CGM in the DiffVAS framework, we assess a DiffVAS variant (denoted Mask-DiffVAS) where we exclude the latent representation of the search space reconstructed using CGM (i.e.~$l_{\mathrm{re}}(t)$, cf.~Fig.~\ref{fig:framework}) from the input state of the $\mathrm{TCPM}$. From Table~\ref{tab: CGM_ablation} we see that DiffVAS significantly outperforms Mask-DiffVAS, with performance increases from 8.1$\%$ to 37.7$\%$. This shows the crucial role of using the latent representation of the synthesized search space $l_{\mathrm{re}}(t)$ for planning and underscores the importance of the CGM within DiffVAS.
\begin{table}[H]
%\vspace{-3mm}
    \centering
    \small
    \caption{Significance of using the latent representation of the conditional generative module (CGM) within DiffVAS.}
    \vspace{-3mm}
    \setlength{\tabcolsep}{2.2pt} % Adjust column spacing for better fit
    \begin{tabular}{p{1.70cm}p{0.78cm}p{0.73cm}p{0.86cm}p{0.73cm}p{0.73cm}p{0.86cm}p{0.73cm}p{0.73cm}p{0.86cm}}
        \toprule
        \multicolumn{4}{c}{Test with $\mathcal{Z}$ = \{ Ship \}} & 
        \multicolumn{3}{c}{Test with $\mathcal{Z}$ = \{ LV \}} & 
        \multicolumn{3}{c}{Test with $\mathcal{Z}$ = \{ Plane \}} \\
        \midrule
        Method & $\mathcal{B}=5$ & $\mathcal{B}=7$ & $\mathcal{B}=10$ & 
        $\mathcal{B}=5$ & $\mathcal{B}=7$ & $\mathcal{B}=10$ & 
        $\mathcal{B}=5$ & $\mathcal{B}=7$ & $\mathcal{B}=10$ \\
        \cmidrule(r){1-4} \cmidrule(r){5-7} \cmidrule(l){8-10} 
        Mask-DiffVAS & 1.82 & 2.65 & 3.29 & 2.32 & 2.91 & 4.95 & 2.45 & 3.23 & 4.03 \\
        \hline 
        \textbf{DiffVAS} & \textbf{2.12} & \textbf{3.22} & \textbf{3.91} & 
        \textbf{2.54} & \textbf{3.57} & \textbf{5.78} & 
        \textbf{3.12} & \textbf{4.07} & \textbf{5.24} \\ 
        \hline
        \hline
        \multicolumn{4}{c}{Test with $\mathcal{Z}$ = \{ Harbor \}} & 
        \multicolumn{3}{c}{Test with $\mathcal{Z}$ = \{ Roundabout \}} & 
        \multicolumn{3}{c}{Test with $\mathcal{Z}$ = \{ Helicopter \}} \\
        \midrule
        Method & $\mathcal{B}=5$ & $\mathcal{B}=7$ & $\mathcal{B}=10$ & 
        $\mathcal{B}=5$ & $\mathcal{B}=7$ & $\mathcal{B}=10$ & 
        $\mathcal{B}=5$ & $\mathcal{B}=7$ & $\mathcal{B}=10$ \\
        \cmidrule(r){1-4} \cmidrule(l){5-7} \cmidrule(l){8-10} 
        Mask-DiffVAS & 1.75 & 2.56 & 3.82 & 1.91 & 2.99 & 4.10 & 1.54 & 3.33 & 4.67 \\
        \hline
        \textbf{DiffVAS} & \textbf{2.01} & \textbf{3.15} & \textbf{4.45} & 
        \textbf{2.32} & \textbf{3.33} & \textbf{4.89} & 
        \textbf{2.12} & \textbf{3.91} & \textbf{5.05} \\ 
        \bottomrule
    \end{tabular}
    \label{tab: CGM_ablation}
    %\vspace{-3mm}
\end{table}
\begin{figure*}
%\vspace{-4mm}
  \centering
  \setlength{\tabcolsep}{1pt}
  \renewcommand{\arraystretch}{1}
  \includegraphics[width=0.96\linewidth]{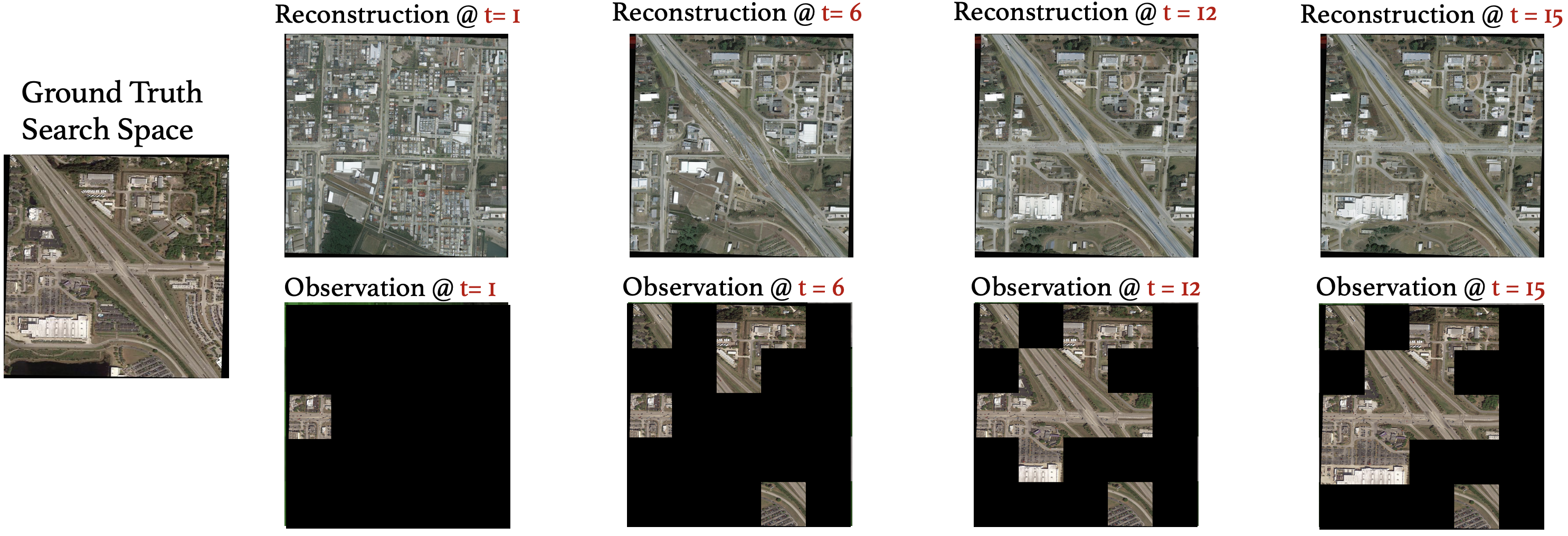}
  \caption{Test set example showing CGM reconstructions from partially observed glimpses at various DiffVAS search stages.}
  \label{fig:vis_tc-povas}
%\vspace{-3mm}
\end{figure*}

\noindent
\textbf{Importance of the $\mathbf{TCPM}$.}
%finetune diffusion with reward 
To assess the importance of the planner module in DiffVAS, we replace the TCPM with a classifier trained to predict a target-containing grid cell based on the same input state $s_t = \left( l^{\mathrm{z}}_{\mathrm{img}}(t),  o^t(\cdot \mid z), \mathcal{B}^{t} \right)$ as the planner.
\begin{table}[t]
%\vspace{-3mm}
    \centering
    \small
    %\vspace{-3mm}
    \caption{The target-conditioned planning module (TCPM) in DiffVAS significantly outperforms a greedy alternative.}
    %\vspace{-3mm}
    \setlength{\tabcolsep}{2.2pt} % Adjust column spacing for better fit
    \begin{tabular}{p{2.02cm}p{0.73cm}p{0.73cm}p{0.86cm}p{0.73cm}p{0.73cm}p{0.86cm}p{0.73cm}p{0.73cm}p{0.86cm}}
        \toprule
        \multicolumn{4}{c}{Test with $\mathcal{Z}$ = \{ Ship \}} & 
        \multicolumn{3}{c}{Test with $\mathcal{Z}$ = \{ LV \}} & 
        \multicolumn{3}{c}{Test with $\mathcal{Z}$ = \{ Plane \}} \\
        \midrule
        Method & $\mathcal{B}=5$ & $\mathcal{B}=7$ & $\mathcal{B}=10$ & 
        $\mathcal{B}=5$ & $\mathcal{B}=7$ & $\mathcal{B}=10$ & 
        $\mathcal{B}=5$ & $\mathcal{B}=7$ & $\mathcal{B}=10$ \\
        \cmidrule(r){1-4} \cmidrule(r){5-7} \cmidrule(l){8-10} 
        Greedy-DiffVAS & 1.29 & 2.01 & 2.96 & 1.81 & 2.45 & 4.46 & 2.00 & 2.57 & 3.77 \\
        \hline 
        \textbf{DiffVAS} & \textbf{2.12} & \textbf{3.22} & \textbf{3.91} & 
        \textbf{2.54} & \textbf{3.57} & \textbf{5.78} & 
        \textbf{3.12} & \textbf{4.07} & \textbf{5.24} \\ 
        \hline
        \hline
        \multicolumn{4}{c}{Test with $\mathcal{Z}$ = \{ Harbor \}} & 
        \multicolumn{3}{c}{Test with $\mathcal{Z}$ = \{ Roundabout \}} & 
        \multicolumn{3}{c}{Test with $\mathcal{Z}$ = \{ Helicopter \}} \\
        \midrule
        Method & $\mathcal{B}=5$ & $\mathcal{B}=7$ & $\mathcal{B}=10$ & 
        $\mathcal{B}=5$ & $\mathcal{B}=7$ & $\mathcal{B}=10$ & 
        $\mathcal{B}=5$ & $\mathcal{B}=7$ & $\mathcal{B}=10$ \\
        \cmidrule(r){1-4} \cmidrule(l){5-7} \cmidrule(l){8-10} 
        Greedy-DiffVAS & 1.23 & 2.19 & 3.32 & 1.22 & 2.57 & 3.92 & 1.11 & 3.02 & 4.34  \\
        \hline
        \textbf{DiffVAS} & \textbf{2.01} & \textbf{3.15} & \textbf{4.45} & 
        \textbf{2.32} & \textbf{3.33} & \textbf{4.89} & 
        \textbf{2.12} & \textbf{3.91} & \textbf{5.05} \\ 
        \bottomrule
    \end{tabular}
    \label{tab: TCPM_ablation}
    %\vspace{-5mm}
\end{table}
The classifier is trained using binary cross-entropy loss. We then compare the performance of this modified version, \emph{Greedy-DiffVAS}, with the original DiffVAS. We emphasize that the only distinction between Greedy-DiffVAS and DiffVAS is the replacement of the planner module with the classifier. We evaluate their performances across different target categories, as reported in Table~\ref{tab: TCPM_ablation}. DiffVAS consistently outperforms Greedy-DiffVAS, with performance increases ranging from 16.4$\%$ to 91.0$\%$ across the various evaluation settings. These empirical results thus demonstrate that relying solely on greedy actions is inadequate for tasks that require a balance between exploration and exploitation, which highlights the critical role of the planning module in learning an efficient search policy in partially observable environments.
\\ \\
\textbf{Impact of $\mathcal{R}^{\mathrm{GR}}$ and $\mathcal{R}^{\mathrm{LU}}$ on search performance.}
%full, only vas, only r_GR, only r_lu, only (r_gr + r_lu).
To assess the significance of various reward components in the reward function (\ref{eq:reward}), we train DiffVAS with different reward components and compare the performances across various target settings. The results in Table~\ref{tab: REWARD_ablation}
\begin{table}
%\vspace{-3mm}
    \centering
    \small
    \caption{Analyzing different components of the proposed reward function. Using the full reward yields the best results.}
    %\vspace{-3mm}
    \setlength{\tabcolsep}{2.2pt} % Adjust column spacing for better fit
    \begin{tabular}{p{1.77cm}p{0.73cm}p{0.76cm}p{0.86cm}p{0.73cm}p{0.73cm}p{0.86cm}p{0.73cm}p{0.73cm}p{0.86cm}}
        \toprule
        \multicolumn{4}{c}{Test with $\mathcal{Z}$ = \{ Ship \}} & 
        \multicolumn{3}{c}{Test with $\mathcal{Z}$ = \{ LV \}} & 
        \multicolumn{3}{c}{Test with $\mathcal{Z}$ = \{ Plane \}} \\
        \midrule
        Reward & $\mathcal{B}=5$ & $\mathcal{B}=7$ & $\mathcal{B}=10$ & 
        $\mathcal{B}=5$ & $\mathcal{B}=7$ & $\mathcal{B}=10$ & 
        $\mathcal{B}=5$ & $\mathcal{B}=7$ & $\mathcal{B}=10$ \\
        \cmidrule(r){1-4} \cmidrule(r){5-7} \cmidrule(l){8-10} 
        $\mathcal{R}^{\mathrm{AS}}$ & 1.65 & 2.71 & 3.77 & 1.89 & 2.85 & 3.90 & 2.05 & 3.50 & 4.68 \\
        $\mathcal{R}^{\mathrm{AS}} + \mathcal{R}^{\mathrm{LU}}$ & 1.71 & 2.79 & 3.79 & 1.90 & 2.92 & 4.11 & 2.09 & 3.53 & 4.74 \\ 
        $\mathcal{R}^{\mathrm{GR}} + \mathcal{R}^{\mathrm{LU}}$ & 1.63 & 2.67 & 3.66 & 1.73 & 2.78 & 3.79 & 1.80 & 3.43 & 4.69 \\
        $\mathcal{R}^{\mathrm{AS}} + \mathcal{R}^{\mathrm{GR}}$ & 1.76 & 2.88 & 3.82 & 1.90 & 2.98 & 4.32 & 1.89 & 3.54 & 4.78 \\
        \hline 
        \textbf{Full reward} & \textbf{2.01} & \textbf{3.15} & \textbf{4.45} & 
        \textbf{2.32} & \textbf{3.33} & \textbf{4.89} & 
        \textbf{2.12} & \textbf{3.91} & \textbf{5.05} \\ 
        \bottomrule
    \end{tabular}
    \label{tab: REWARD_ablation}
    %\vspace{-5mm}
\end{table}
suggest that relying only on $\mathcal{R}^{\mathrm{AS}}$ is insufficient, which shows the importance of actions that enhance information gathering about the search space. However, as would be expected, merely gathering information is not enough, as performance drops when training the policy using only $\mathcal{R}^{\mathrm{GR}} + \mathcal{R}^{\mathrm{LU}}$. Thus, incorporating both $\mathcal{R}^{\mathrm{AS}}$ and $\mathcal{R}^{\mathrm{GR}} + \mathcal{R}^{\mathrm{LU}}$ is essential for learning an effective search policy in partially observed environments. Additionally, we observe a performance drop (line 4) when we exclude $\mathcal{R}^{\mathrm{LU}}$ from the full reward (\ref{eq:reward}),
%Furthermore, we observe a slight improvement in search performance when incorporating $R^{LU}$ into the reward function (see rows 1 and 2) for training the TCPM module,
which shows the importance of the local uncertainty-based reward $R^{LU}$.
\\ \\
\textbf{Effectiveness in handling multiple target categories.}
% simultaneously}
% pass all the target category to clip ...get all the embedding from clip individually and average them before applying cross attention... directly concat the prompt without cross attention...
\begin{table}[b]
%\vspace{-3mm}
    \centering
    \small
    %\vspace{-2mm}
    \caption{DiffVAS uses the most effective inference strategy.}
    %\vspace{-3mm}
    \setlength{\tabcolsep}{2.2pt} % Adjust column spacing for better fit
    \begin{tabular}{p{1.61cm}p{0.73cm}p{0.73cm}p{0.86cm}p{0.73cm}p{0.73cm}p{0.86cm}p{0.73cm}p{0.73cm}p{0.86cm}}
        \toprule
        \multicolumn{4}{c}{Test with $\mathcal{Z}$ = \{ Ship, Harbor \}} & 
        \multicolumn{3}{c}{Test with $\mathcal{Z}$ = \{ LV, SV \}} & 
        \multicolumn{3}{c}{Test with $\mathcal{Z}$ = \{ Plane, Heli \}} \\
        \midrule
        Method & $\mathcal{B}=5$ & $\mathcal{B}=7$ & $\mathcal{B}=10$ & 
        $\mathcal{B}=5$ & $\mathcal{B}=7$ & $\mathcal{B}=10$ & 
        $\mathcal{B}=5$ & $\mathcal{B}=7$ & $\mathcal{B}=10$ \\
        \cmidrule(r){1-4} \cmidrule(r){5-7} \cmidrule(l){8-10} 
        Avg-DiffVAS & 2.45 & 3.32 & 4.45 & 2.51 & 3.82 & 5.10 & 2.21 & 4.09 & 5.55 \\
        Emb-DiffVAS & 2.67 & 3.55 & 4.67 & 2.81 & 4.02 & 5.31 & 2.45 & 4.23 & 5.89 \\
        \hline 
        \textbf{DiffVAS} & \textbf{2.98} & \textbf{4.16} & \textbf{4.92} & 
        \textbf{3.05} & \textbf{4.33} & \textbf{5.52} & 
        \textbf{3.11} & \textbf{4.34} & \textbf{6.02} \\
        \bottomrule
    \end{tabular}
    \label{tab: inference}
    %\vspace{-4mm}
\end{table}
We evaluate
%assess the efficacy of 
the proposed inference approach (see Sec.~\ref{sec:inf}) by comparing DiffVAS with two variants that use the same training strategy, but differ during inference in the way $l_{\mathrm{z}}$ is computed: (1) \emph{Avg-DiffVAS} computes $l_{\mathrm{z}}$ by inputting the entire target category set $\mathcal{Z}$ into the CLIP text encoder, requiring only a single forward pass through the planning module at each time step,
%, unlike our proposed inference approach;
and (2) \emph{Emb-DiffVAS} computes target-specific embeddings by processing each target category in the set $\mathcal{Z}$ individually through the CLIP text encoder, then averages them to obtain $l_{\mathrm{z}}$. We compare their performances across different $\mathcal{Z}$ in Table~\ref{tab: inference} and observe that these alternative strategies perform worse than our proposed strategy.
\\ \\
\textbf{Visualizing reconstructions from the CGM.}
% for same image same budget but vary set of target object and how exploration changes..
%Figure~\ref{fig:vis_tc-povas} illustrates an example of DiffVAS exploration behavior in a search space with different target sets $\mathcal{Z}$. Additional visualizations are in the Appendix.
Fig.~\ref{fig:vis_tc-povas} illustrates an example of the CGM's reconstruction of the search space from partially observed glimpses.
%; see more in the appendix.
% \begin{figure*}
% \vspace{-4mm}
%   \centering
%   \setlength{\tabcolsep}{1pt}
%   \renewcommand{\arraystretch}{1}
%   \includegraphics[width=\linewidth]{ICLR 2025 Template/figures/cgm_1.png}
%   \caption{CGM’s reconstruction from partially observed glimpses at various search stages.}% under uniform cost.}
%   \label{fig:vis_tc-povas}
% \vspace{-3mm}
% \end{figure*}
\\ \\
\textbf{Zero-shot generalization.}
% evaluate DiffVAS with target category that is unseen during training phase.
To assess the zero-shot generalizability of DiffVAS, we evaluate on xView a policy trained solely on DOTA, while ensuring that the target category set $\mathcal{Z}$ from DOTA differs from that in xView (this has to be done, since the categories partially overlap between these datasets). The results in Table~\ref{tab: zero-shot} show performance improvements ranging between 36.3\% to 281.5\% compared to the baseline approaches and highlight the effectiveness of DiffVAS in zero-shot generalization. The superior zero-shot generalizability of DiffVAS stems from the CGM module, which preserves the strength of the trained diffusion model. This ensures that the representation extracted from CGM (i.e.~$l_{\mathrm{re}}(t)$, $l_{\mathrm{h}}(t)$), a key component of the planning module's state input ($s_t$), remains robust. %See the appendix for additional results.
\begin{table}[t]
%\vspace{-2mm}
    \centering
    \small
    \caption{DiffVAS demonstrates superior zero-shot generalization (here: DOTA to xView) compared to alternative methods.}
    \vspace{-3mm}
    \setlength{\tabcolsep}{2.2pt} % Adjust column spacing for better fit
    \begin{tabular}{p{1.5cm}p{0.73cm}p{0.73cm}p{0.86cm}p{0.73cm}p{0.73cm}p{0.86cm}p{0.73cm}p{0.73cm}p{0.86cm}}
        \toprule
        \multicolumn{4}{c}{\>\>\>Test with $\mathcal{Z}$ = \{ Small Car \}} & 
        \multicolumn{3}{c}{Test with $\mathcal{Z}$ = \{ Sail Boat \}} & 
        \multicolumn{3}{c}{Test with $\mathcal{Z}$ = \{ Helipad \}} \\
        \midrule
        Method & $\mathcal{B}=5$ & $\mathcal{B}=7$ & $\mathcal{B}=10$ & 
        $\mathcal{B}=5$ & $\mathcal{B}=7$ & $\mathcal{B}=10$ & 
        $\mathcal{B}=5$ & $\mathcal{B}=7$ & $\mathcal{B}=10$ \\
        \cmidrule(r){1-4} \cmidrule(r){5-7} \cmidrule(l){8-10} 
        E2EVAS & 1.51 & 2.03 & 3.04 & 0.25 & 0.35 & 0.47 & 0.15 & 0.21 & 0.29 \\
        MPS-VAS & 1.54 & 2.09 & 3.12 & 0.27 & 0.36 & 0.49 & 0.16 & 0.31 & 0.38 \\
        \hline 
        \textbf{DiffVAS} & \textbf{2.10} & \textbf{2.95} & \textbf{4.34} & 
        \textbf{1.03} & \textbf{1.19} & \textbf{1.30} & 
        \textbf{0.45} & \textbf{0.89} & \textbf{1.02} \\ 
        \bottomrule
    \end{tabular}
    \label{tab: zero-shot}
    \vspace{-6mm}
\end{table}
\subsection{Related Work}
Our work bridges and expands concepts from visual active search, autonomous UAVs, and active scene reconstruction; we next briefly mention relevant prior works within each broad category.
\\ \\
\textbf{Visual active search (VAS).} The VAS framework was first introduced in \cite{sarkar2024visual}, who framed it as a budget-constrained MDP and tackled it using deep RL. In \cite{sarkar2023partially,sarkar2024geospatial}, a meta-learning approach is introduced that enables the policy to use supervised information gathered during the search. Key limitations of prior works are the reliance on full observations of search areas and the focus on target-specific policies, which makes them incapable of handling multiple target categories simultaneously. Active geo-localization~\citep{pirinen2022aerial,sarkar2024gomaa} is a task similar to VAS, in which an agent with aerial view observations of a scene seeks to actively localize a goal. However, this task considers only the single-target location and assumes access to an observation of the target location.  %consider only static objects (e.g.~salient buildings).
\\ \\
\textbf{Autonomous UAV exploration.} Our work falls within the broad literature on autonomous control and navigation of UAVs~\citep{area2,popovic2020informative,stache2022adaptive,area1,urbanmapping,astar,sadat2015fractal}. Many of these prior works~\citep{wu2019liteeval,yang2020resolution,wang2020object,thavamani2021fovea,meng2022count,meng2022adavit} 
assume access to a global lower-resolution observation of the whole area of interest a priori, while 
DiffVAS \emph{reconstructs} the area from partial observations on the fly.
\\ \\
\textbf{Active scene/object reconstruction.} There is extensive prior work on active reconstruction of scenes and/or objects~\citep{jayaraman2016look, jayaraman2018learning, xiong2018snap, pirinen2019domes}. These methods typically focus solely on optimizing for reconstruction, while our goal is identifying target-rich regions. Our task setup demands balancing \emph{exploration} (obtaining useful information about the scene) and \emph{exploitation} (finding target objects).
%THIS I COPIED FROM ICLR REBUTTAL, BUT I DONT THINS WE NEED TO CITE THESE WORKS, THEY ARE NOT ABOUT ACTIVE RECONSTRUCTION / ALEKSIS. The problem we are tackling in this work are fundamentally different than the mentioned work, such as MDT (ICCV23) and DiffMAE (ICCV23). Both these works aim to reconstruct an image from the partially observed scene, while we tackle a more challenging problem of active discovery of target objects. Hence, these methods (such as MDT, and DiffMAE) typically focus solely on optimizing for reconstruction, while our ultimate goal is identifying target-rich regions. Success for our task hinges on balancing exploration (obtaining useful information about the scene) and exploitation (finding objects of interest). To justify why such a reconstruction-only method is not appropriate to tackle active search problems, we performed an experiment with Greedy-DiffVAS that utilized the reconstructed features only to decide the next query location and compared the performance with our proposed framework. The empirical outcomes highlight the critical role of the planning module (TCPM) in learning an efficient search policy in partially observable environments. Please refer to lines 441- 460 in the manuscript for a discussion of this.
%\vspace{-3mm}
\section{Conclusions}
We have presented DiffVAS, a novel multi-target visual active search approach that generalizes across domains. At its core is a diffusion-based conditional generative module that dynamically reconstructs the search area, which enables a target-conditioned planning module to plan movements effectively in a partially observable environment. Furthermore, our inference method enables DiffVAS to handle tasks that involve searching multiple target categories simultaneously, with varying category counts. Trained with a novel reward balancing exploration and exploitation, DiffVAS outperforms strong baselines and prior methods, while demonstrating excellent zero-shot generalization. We hope our proposed framework will prove useful in a variety of practical scenarios, spanning from search-and-rescue operations to combating human trafficking. %Trained models and code are available at this \href{https://github.com/mvrl/Multi-objective-active-sampling-for-inpainting}{\textcolor{blue}{link}}.

\subsection*{Acknowledgments}

This work was partially supported by the NSF (IIS-2214141), ARO (W911NF-25-1-0059), ONR (N000142412663), Foresight Institute, and Amazon.

\bibliographystyle{ACM-Reference-Format} 
\bibliography{sample}

\newpage

\clearpage
\section*{Appendix}
In this appendix, we provide several additional quantitative and qualitative results, including various ablation studies. We list the content of the appendix below:
\begin{itemize}[noitemsep,topsep=0pt, leftmargin=*]
    \item In Appendix~\ref{app:eval_xview}, we present the results with the xView dataset, including both single- and multi-target category search settings.
    \item In Appendix~\ref{app:grid_size}, we compare the performance of \emph{DiffVAS} with baselines across different grid sizes. 
    \item Additional visualizations of the CGM's search space reconstruction from partially observed glimpses are presented in Appendix~\ref{app:cgm_vis}.
    \item The implementation details of different modules of our proposed \emph{DiffVAS} framework is presented in Appendix~\ref{app:imple}. In this section we also provide the anonymous code link for training and inference of \emph{DiffVAS}, and include a discussion on computing resources.
    \item We present additional visualizations of the exploration behavior of \emph{DiffVAS} in Appendix~\ref{app:exp_vis}.
    \item We analyze the effectiveness of the cross-attention layer in Appendix~\ref{app:cross_att}.
    \item We compare the performance of \emph{DiffVAS} with a similar approach that assumes full observability of the search space, and report our findings in Appendix~\ref{app:full}.
    \item In Appendix~\ref{app:more_vis}, we present more visualizations of CGM's search space reconstructions from partially observed glimpses.
    \item In Appendix~\ref{app:select}, we detail the reason for selecting the target categories for evaluation. 
    \item The procedure for sampling an episode during training is discussed in Appendix~\ref{app:sampling}.
    \item We explore potential future directions for the proposed TC-POVAS task, inspired by real-world challenges, in Appendix~\ref{app:FD}. 
    \item Further details of the cross-attention layer are discussed in Appendix~\ref{app:cross}.
    \item Finally, we discuss the impact statement in Appendix~\ref{app:is}.
\end{itemize}

\begin{comment}
% Table of Contents
%\frontmatter  % Ensures Roman numerals for ToC and Preface pages
%\setcounter{tocdepth}{0}
\tableofcontents  % Generates Table of Contents

%\mainmatter  % Switches to Arabic numbering for main chapters

% Chapters
%\chapter{Introduction}
%This is the introduction.

%\chapter{Literature Review}
\section{Evaluation with xView Dataset}
\subsection{Evaluation in single-target category search tasks}
\subsection{Evaluation in multi-target category search tasks}

%\chapter{Methodology}
\section{Evaluation with Different Grid Sizes}
%Explanation of the process.

%\chapter{Results}
\section{Visualizations of CGM search space reconstructions from partially observed glimpses}

\section{Implementations Details}
\subsection{Details of CGM}
\subsection{Details of TCPM}
\subsection{Compute resources}

\section{Visualization of Exploration behavior of Diff-VAS}

\section{Effectiveness of Cross-attention Layer}
\section{Comparison with fully observable search setting}
\section{More Visualizations of CGM Search Space Reconstructions from Partially Observed Glimpses}
\section{Reasoning for Selecting the Target Categories for Evaluation}
\section{Procedure for Sampling an Episode During Training}
\section{Efficacy of Local Uncertainty-Based Reward}
\section{Further Details of Cross-Attention Layer}
\end{comment}

%%%%%%%%%%%%%%%%%%%%%%%%%
%\chapter{Conclusion}
%Final thoughts.

\subsection{Evaluation with xView Dataset} \label{app:eval_xview}
%both single and multi target setting
\paragraph{Evaluation in single-target category search tasks}
In this section we compare DiffVAS to baseline approaches using the xView dataset, starting with a single-target category search setting. The evaluation includes target classes such as Small Car (SC), Helicopter, Sail Boat (SB), Container Ship, Building, and Helipad. The results in Table~\ref{tab: xview_single} reveal a similar trend to those observed with the DOTA dataset (main paper), with significant performance improvements of the proposed DiffVAS approach over all baselines, ranging from 11.2\% to 109.7\% compared to the strongest baseline, MPS-VAS. The empirical results further confirm the effectiveness of our proposed DiffVAS framework in learning an efficient visual active search policy in partially observed environments.

\paragraph{Evaluation in multi-target category search tasks}
In this section we evaluate DiffVAS using the xView dataset with  $\mathcal{Z}$ containing multiple target categories, and the results are presented in Table~\ref{tab: xview_multi}. Consistent with our findings on the DOTA dataset (main paper), we observe a notable improvement in performance across different target category sets, ranging from 8.8\% to 17.3\% compared to the strongest baseline, MPS-VAS. This further underscores the effectiveness of our proposed DiffVAS inference strategy in handling diverse and complex search tasks involving multiple target categories.
\begin{table}[H]
    \centering
    \small
    \caption{ANT comparisons on the xView dataset for the single-target category setting.}
    %\vspace{-2mm}
    \setlength{\tabcolsep}{2.5pt} 
    \begin{tabular}{p{1.4cm}p{0.73cm}p{0.73cm}p{0.86cm}p{0.73cm}p{0.73cm}p{0.86cm}p{0.73cm}p{0.73cm}p{0.86cm}}
        \toprule
        \multicolumn{4}{c}{\>\>\>Test with $\mathcal{Z}$ = \{ SC \}} & \multicolumn{3}{c}{Test with $\mathcal{Z}$ = \{ Helicopter \}} & \multicolumn{3}{c}{Test with $\mathcal{Z}$ = \{ Bus \}}\\
        \midrule
        Method & $\mathcal{B}=5$ & $\mathcal{B}=7$ & $\mathcal{B}=10$ & $\mathcal{B}=5$ & $\mathcal{B}=7$ & $\mathcal{B}=10$ & $\mathcal{B}=5$ & $\mathcal{B}=7$ & $\mathcal{B}=10$ \\
        \cmidrule(r){1-4} \cmidrule(r){5-7} \cmidrule(l){8-10}
        RS & 1.92 & 2.51 & 3.51 & 0.17 & 0.24 & 0.41 & 0.31 & 0.35 & 0.48 \\
        E2EVAS & 2.37 & 3.07 & 3.88 & 0.19 & 0.28 & 0.40 & 0.29 & 0.41 & 0.47 \\ 
        MPS-VAS & 2.45 & 3.12 & 3.93 & 0.23 & 0.31 & 0.43 & 0.30 & 0.43 & 0.51 \\ 
        \hline 
        \textbf{DiffVAS} & \textbf{2.91} & \textbf{3.89} & \textbf{4.53} & \textbf{0.45} & \textbf{0.65} & \textbf{0.81} & \textbf{0.42} & \textbf{0.56} & \textbf{0.66} \\ 
        \hline
        \hline
        \multicolumn{4}{c}{Test with $\mathcal{Z}$ = \{ Building \}} & \multicolumn{3}{c}{Test with $\mathcal{Z}$ = \{ Container Ship \}} & \multicolumn{3}{c}{Test with $\mathcal{Z}$ = \{ Truck \}}\\
        \midrule
        Method & $\mathcal{B}=5$ & $\mathcal{B}=7$ & $\mathcal{B}=10$ & $\mathcal{B}=5$ & $\mathcal{B}=7$ & $\mathcal{B}=10$ & $\mathcal{B}=5$ & $\mathcal{B}=7$ & $\mathcal{B}=10$ \\
        \cmidrule(r){1-4} \cmidrule(l){5-7} \cmidrule(l){8-10}
        RS & 2.34 & 3.32 & 3.91 & 0.20 & 0.31 & 0.42 &0.18 & 0.31 & 0.40 \\
        E2EVAS & 2.61 & 3.45 & 4.18 & 0.21 & 0.35 & 0.47 & 0.18 & 0.33 & 0.42 \\ 
        MPS-VAS & 2.68 & 3.51 & 4.22 & 0.23 & 0.36 & 0.48 & 0.21 & 0.37 & 0.45 \\
        \hline
        \textbf{DiffVAS} & \textbf{2.93} & \textbf{3.92} & \textbf{4.52} & \textbf{0.31} & \textbf{0.45} & \textbf{0.60} & \textbf{0.29} & \textbf{0.50} & \textbf{0.62} \\ 
        \bottomrule
    \end{tabular}
    \label{tab: xview_single}
    %\vspace{-4mm}
\end{table}

\begin{table}[H]
    \centering
    %\small
    %\footnotesize
    \small
    %\scriptsize
    \caption{ANT comparisons on the xView dataset for the multiple-target category setting.}
    %\vspace{-2mm}
    \setlength{\tabcolsep}{2.2pt} 
    \begin{tabular}{p{1.90cm}p{0.73cm}p{0.73cm}p{0.86cm}p{0.73cm}p{0.73cm}p{0.86cm}p{0.73cm}p{0.73cm}p{0.86cm}}
        \toprule
        \multicolumn{4}{c}{\>\>\>Test with $\mathcal{Z}$ = \{ SC, Bus \}} & \multicolumn{3}{c}{Test with $\mathcal{Z}$ = \{ SC, Truck \}} & \multicolumn{3}{c}{Test with $\mathcal{Z}$ = \{ SC, Building \}}\\
        \midrule
        Method & $\mathcal{B}=5$ & $\mathcal{B}=7$ & $\mathcal{B}=10$ & $\mathcal{B}=5$ & $\mathcal{B}=7$ & $\mathcal{B}=10$ & $\mathcal{B}=5$ & $\mathcal{B}=7$ & $\mathcal{B}=10$ \\
        %\midrule
        \cmidrule(r){1-4} \cmidrule(r){5-7} \cmidrule(l){8-10} % For better column separation, thanks to Mico!
        RS & 2.12 & 2.67 & 3.31 & 2.01 & 2.89 & 3.77 & 2.53 & 3.91&5.44 \\
        E2EVAS & 2.46 & 3.25 & 4.05 & 2.43 & 3.17 & 3.96 & 2.97 & 4.66 & 5.31 \\ 
        MPS-VAS & 2.53 & 3.32 & 4.11 & 2.48 & 3.25 & 4.03 & 3.08 & 4.95 & 5.45 \\ 
        \hline 
        \textbf{DiffVAS} & \textbf{3.02} & \textbf{3.81} & \textbf{4.54} & \textbf{2.91} & \textbf{3.99} & \textbf{4.46} & \textbf{3.67} & \textbf{5.32} & \textbf{5.93} \\ 
        % \hline
        % \hline
        % \multicolumn{4}{c}{Test with \mathcal${Z}$ = \{ A,B \}} & \multicolumn{3}{c}{Test with \mathcal${Z}$ = \{ C,D \}} & \multicolumn{3}{c}{Test with \mathcal${Z}$ = \{ E,F \}}\\
        % \midrule
        % Method & $\mathcal{B}=5$ & $\mathcal{B}=7$ & $\mathcal{B}=10$ & $\mathcal{B}=5$ & $\mathcal{B}=7$ & $\mathcal{B}=10$ & $\mathcal{B}=5$ & $\mathcal{B}=7$ & $\mathcal{B}=10$ \\
        % %\midrule
        % \cmidrule(r){1-4} \cmidrule(l){5-7} \cmidrule(l){8-10} % For better column separation, thanks to Mico!
        % RS &  &  &  &  &  &  &  &  &  \\
        % E2EVAS &  &  &  &  &  &  &  &  &  \\ 
        % MPS-VAS &  &  &  &  &  &  &  &  & \\
        % \hline
        % \textbf{DiffVAS} & \textbf{TODO} & \textbf{TODO} & \textbf{TODO} & \textbf{TODO} & \textbf{TODO} & \textbf{TODO} & \textbf{TODO} & \textbf{TODO} & \textbf{TODO} \\ 
        \bottomrule
    \end{tabular}
    \label{tab: xview_multi}
    %\vspace{-4mm}
\end{table}

\subsection{Evaluation with Different Grid Sizes}\label{app:grid_size}
%will go in supplement evaluate with 10 * 10 settings.
We evaluate DiffVAS performance on a 10 $\times$ 10 grid using the DOTA dataset, with results shown in Table~\ref{tab: grid_size1}. Table~\ref{tab: grid_size1} shows results for the single-target category search task.
We compare the search performance of the proposed approach against the baselines across different search budgets, $\mathcal{B} \in \{25, 30, 35\}$, on a 10 $\times$ 10 grid.
Similar to the 5 $\times$ 5 setting, we observe a performance improvement over the baselines, ranging from 0.1$\%$ to 8.8$\%$ across different evaluation scenarios. As anticipated, the overall performance is lower in the larger grid setting, highlighting the increased difficulty and motivating further research into VAS in partially observable environments.
\begin{table}[H]
    \centering
    %\small
    %\footnotesize
    \small
    %\scriptsize
    \caption{ANT comparisons on the DOTA dataset for the single-target category in 10 $\times$ 10 settings.}
    %\vspace{-2mm}
    \setlength{\tabcolsep}{2.2pt} % Adjust column spacing for better fit
    \begin{tabular}{p{2.04cm}p{0.88cm}p{0.88cm}p{0.99cm}p{0.93cm}p{0.93cm}p{1.03cm}p{0.98cm}p{0.83cm}p{0.93cm}}
        \toprule
        \multicolumn{4}{c}{\>\>\>Test with $\mathcal{Z}$ = \{ Large Vehicle \}} & \multicolumn{3}{c}{Test with $\mathcal{Z}$ = \{ Helicopter \}} & \multicolumn{3}{c}{Test with $\mathcal{Z}$ = \{ Plane \}}\\
        \midrule
        Method & $\mathcal{B}=25$ & $=30$ & $=35$ & $\mathcal{B}=25$ & =30 & =35 & $\mathcal{B}=25$ & =30 & =35 \\
        %\midrule
        \cmidrule(r){1-4} \cmidrule(r){5-7} \cmidrule(l){8-10} % For better column separation, thanks to Mico!
        RS & 5.92 &  7.12 & 8.21 & 1.19 & 1.32 & 1.41 & 5.32 & 7.02 & 8.19 \\
        E2EVAS & 6.34 & 7.74 & 8.91 & 1.27 & 1.44 & 1.56 & 5.87 & 7.74 & 8.92 \\ 
        MPS-VAS & 6.37 & 7.80 & 8.96 & 1.32 & 1.49 &   1.60 & 5.93 & 7.83 & 9.01 \\ 
        \hline 
        \textbf{DiffVAS} & \textbf{6.39} & \textbf{7.94} & \textbf{9.02} & \textbf{1.39} & \textbf{1.55} & \textbf{1.74} & \textbf{6.12} & \textbf{7.99} & \textbf{9.21} \\ 
        \hline
        \hline
        \multicolumn{4}{c}{Test with $\mathcal{Z}$ = \{ Roundabout \}} & \multicolumn{3}{c}{Test with $\mathcal{Z}$ = \{ Ship \}} & \multicolumn{3}{c}{Test with $\mathcal{Z}$ = \{ Harbor \}}\\
        \midrule
        Method & $\mathcal{B}=25$ & =30 & =35 & $\mathcal{B}=25$ & =30 & =35 & $\mathcal{B}=25$ & =30 & =35 \\
        %\midrule
        \cmidrule(r){1-4} \cmidrule(l){5-7} \cmidrule(l){8-10} % For better column separation, thanks to Mico!
        RS & 5.11 & 7.02 & 8.13 & 1.10 & 1.15 & 1.47 & 5.47 & 7.02 & 8.57 \\
        E2EVAS & 5.75 & 7.63 & 8.69 & 4.78 & 5.96 & 7.88 & 6.15 & 7.98 & 9.24 \\ 
        MPS-VAS & 5.82 & 7.71 & 8.78 & 4.83 & 6.04 & 7.97 & 6.19 & 8.03 & 9.35 \\
        \hline
        \textbf{DiffVAS} & \textbf{5.94} & \textbf{7.92} & \textbf{8.89} & \textbf{5.05} & \textbf{6.23} & \textbf{8.09} & \textbf{6.31} & \textbf{8.32} & \textbf{9.56} \\ 
        \bottomrule
    \end{tabular}
    \label{tab: grid_size1}
    %\vspace{-4mm}
\end{table}

% \begin{table}[H]
%     \centering
%     %\small
%     %\footnotesize
%     %\tiny
%     \scriptsize
%     \caption{\textbf{ANT} comparisons on the DOTA dataset for the multiple-target category in 10 $\times$ 10 settings.}
%     \vspace{-2mm}
%     \begin{tabular}{p{1.5cm}p{0.93cm}p{0.93cm}p{0.93cm}p{0.93cm}p{0.93cm}p{0.93cm}p{0.93cm}p{0.93cm}p{0.90cm}}
%         \toprule
%         \multicolumn{4}{c}{\>\>\>Test with $\mathcal{Z}$ = \{ X,Y \}} & \multicolumn{3}{c}{Test with $\mathcal{Z}$ = \{ Q,W \}} & \multicolumn{3}{c}{Test with $\mathcal{Z}$ = \{ R,T\}}\\
%         \midrule
%         Method & $\mathcal{B}=25$ & $\mathcal{B}=30$ & $\mathcal{B}=35$ & $\mathcal{B}=25$ & $\mathcal{B}=30$ & $\mathcal{B}=35$ & $\mathcal{B}=25$ & $\mathcal{B}=30$ & $\mathcal{B}=35$ \\
%         %\midrule
%         \cmidrule(r){1-4} \cmidrule(r){5-7} \cmidrule(l){8-10} % For better column separation, thanks to Mico!
%         RS &  &  &  &  &  &  &  &  &  \\
%         E2EVAS &  &  &  &  &  &  &  &  &  \\ 
%         MPS-VAS &  &  &  &  &  &  &  &  &  \\ 
%         \hline 
%         \textbf{DiffVAS} & \textbf{TODO} & \textbf{TODO} & \textbf{TODO} & \textbf{TODO} & \textbf{TODO} & \textbf{TODO} & \textbf{TODO} & \textbf{TODO} & \textbf{TODO} \\ 
%         \bottomrule
%     \end{tabular}
%     \label{tab: grid_size2}
%     %\vspace{-4mm}
% \end{table}

\subsection{Visualizations of CGM search space reconstructions from partially observed glimpses}\label{app:cgm_vis}
\begin{figure*}
%\vspace{-4mm}
  \centering
  \setlength{\tabcolsep}{1pt}
  \renewcommand{\arraystretch}{1}
  \includegraphics[width=1.0\linewidth]{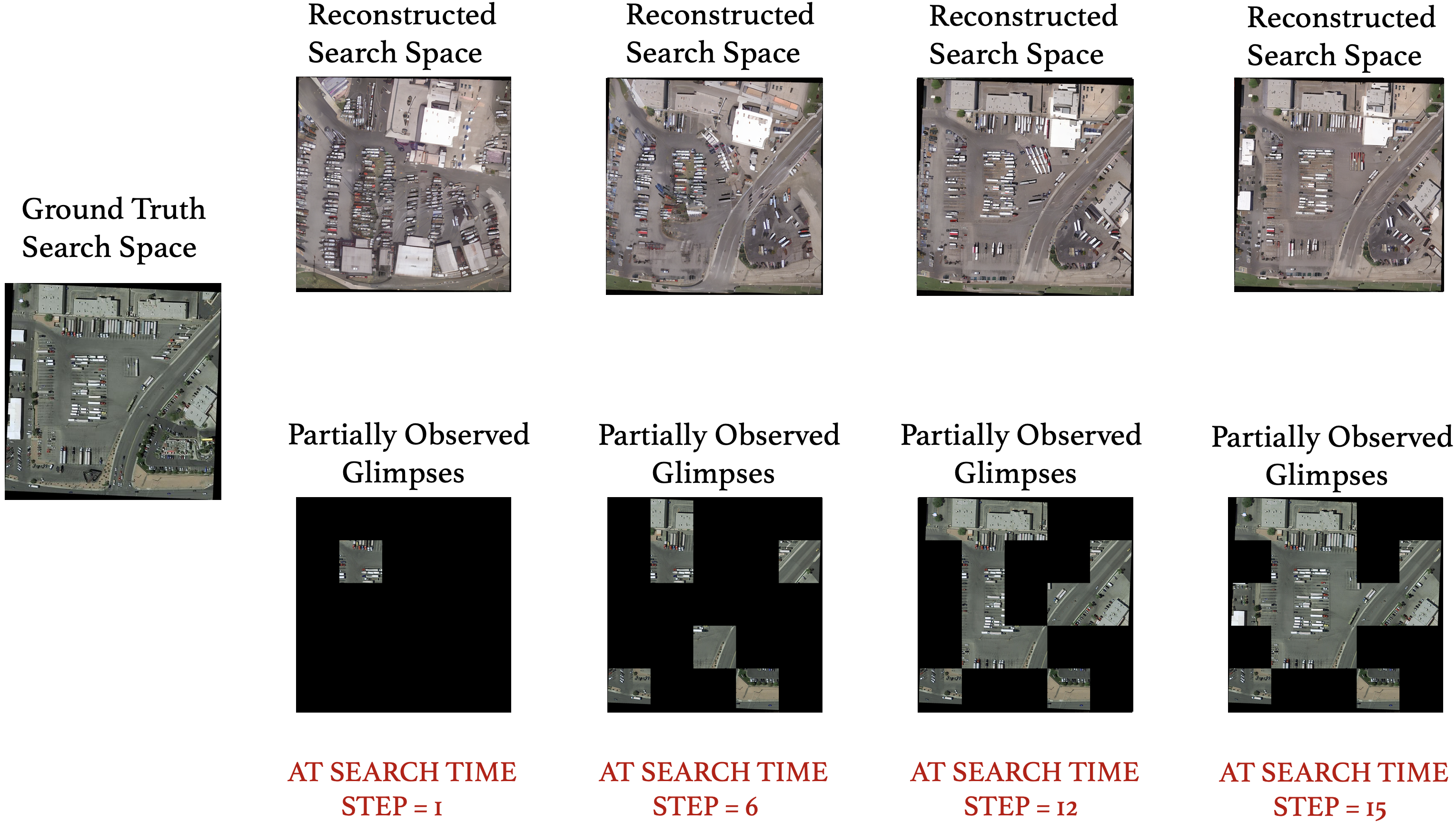}
  \caption{Visualizations of CGM’s reconstruction of the search space from partially observed glimpses at various stages of the search. The reconstruction quality improves as more patches are revealed.}% under uniform cost.}
  \label{fig:cgm_vis1}
%\vspace{-3mm}
\end{figure*}

\begin{figure*}
%\vspace{-4mm}
  \centering
  \setlength{\tabcolsep}{1pt}
  \renewcommand{\arraystretch}{1}
  \includegraphics[width=1.0\linewidth]{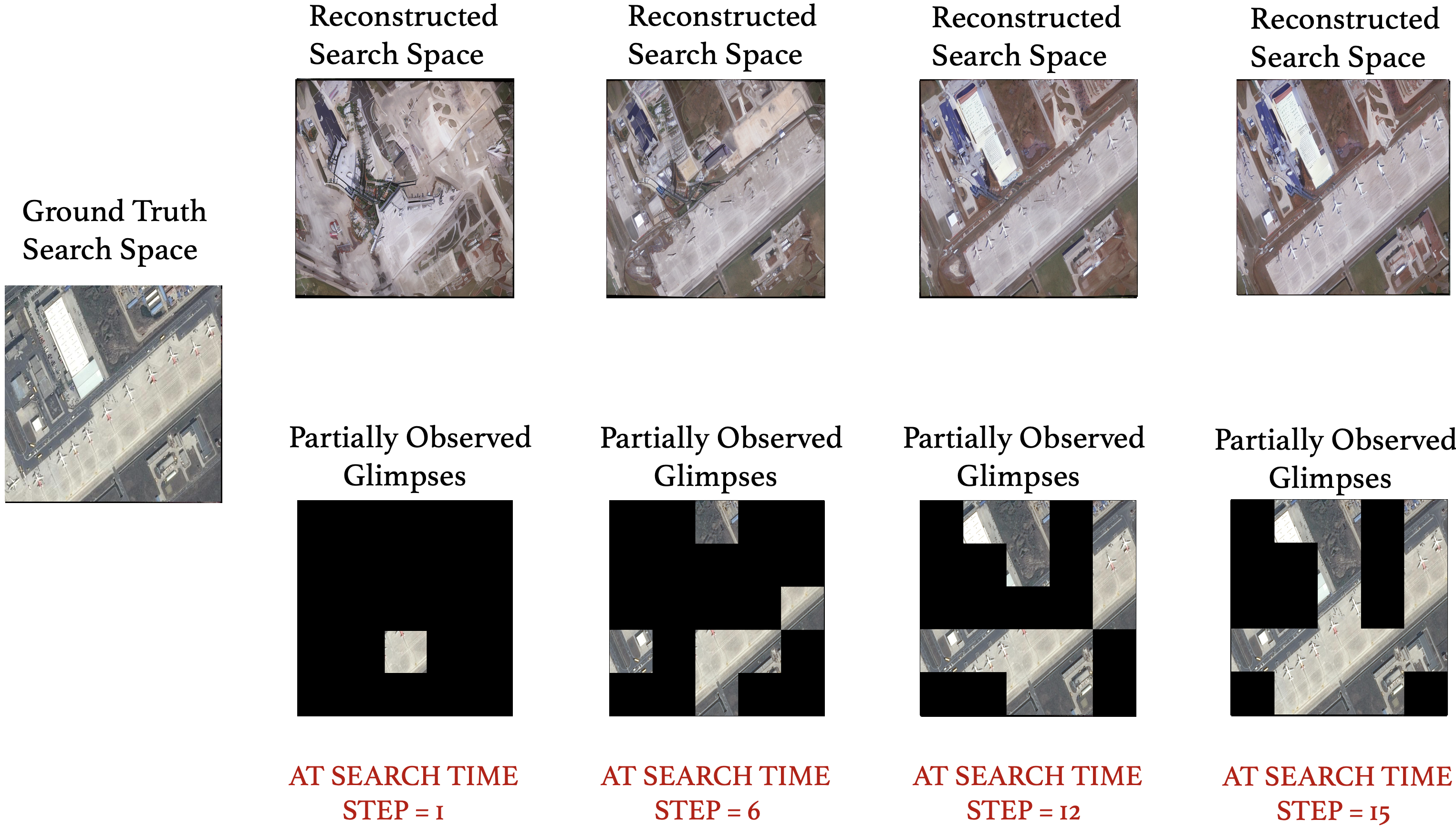}
  \caption{Visualizations of CGM’s reconstruction of the search space from partially observed glimpses at various stages of the search. The reconstruction quality improves as more patches are revealed.}% under uniform cost.}
  \label{fig:cgm_vis2}
%\vspace{-3mm}
\end{figure*}

\begin{figure*}
%\vspace{-4mm}
  \centering
  \setlength{\tabcolsep}{1pt}
  \renewcommand{\arraystretch}{1}
  \includegraphics[width=1.0\linewidth]{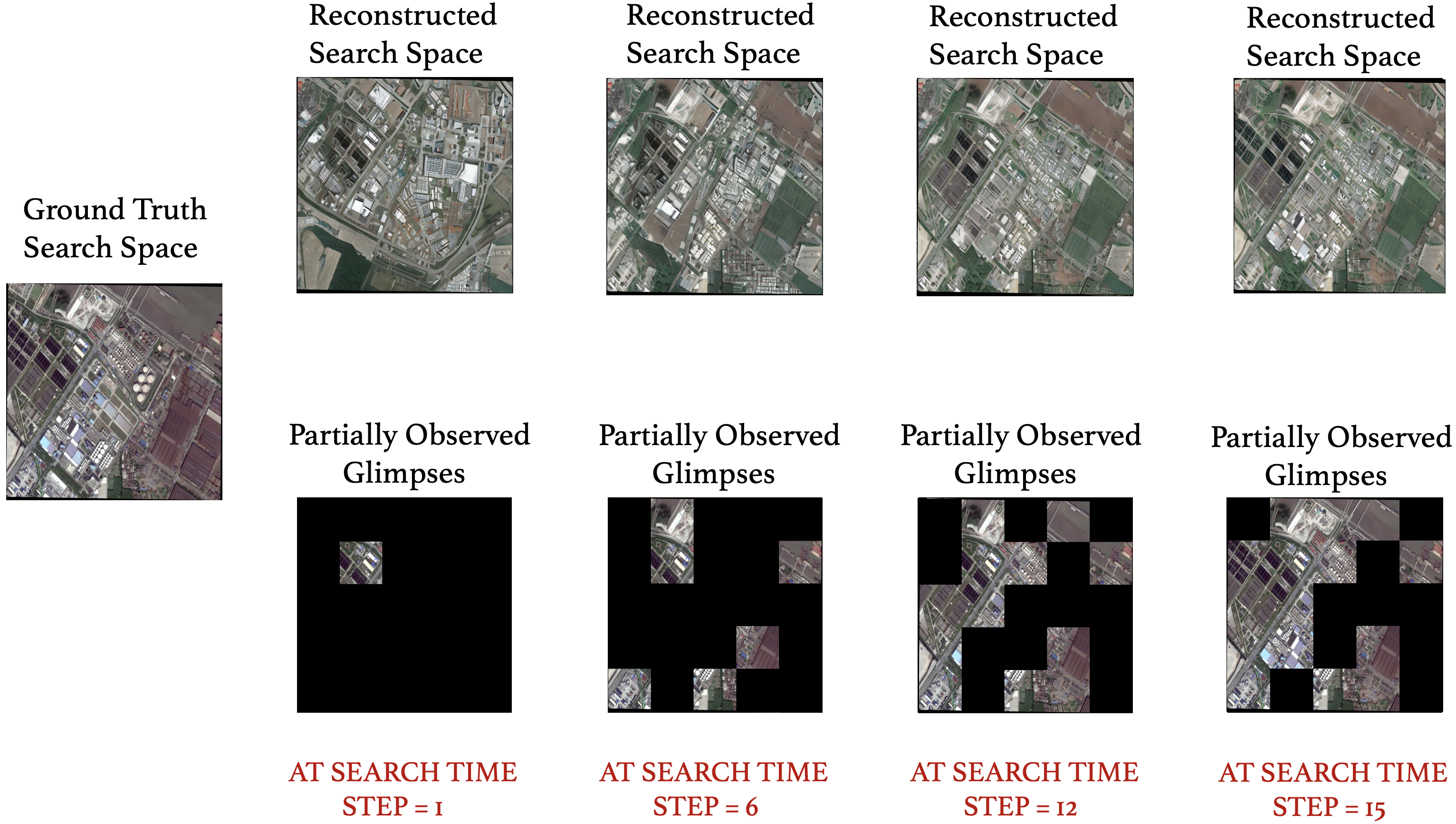}
  \caption{Visualizations of CGM’s reconstruction of the search space from partially observed glimpses at various stages of the search. The reconstruction quality improves as more patches are revealed.}% under uniform cost.}
  \label{fig:cgm_vis3}
%\vspace{-3mm}
\end{figure*}

In this section we provide additional illustrative visualizations of CGM’s reconstruction of search spaces from partially observed glimpses at various stages of the search, corresponding to different history lengths ($h_t$). These visualizations are obtained using the CGM trained with the DOTA dataset. We depict the visualizations in Fig.~\ref{fig:cgm_vis1}, \ref{fig:cgm_vis2} and \ref{fig:cgm_vis3}. These visualizations offer a qualitative perspective on CGM's search space reconstruction quality derived from partially observed glimpses.

\subsection{Implementation Details}\label{app:imple}
In this section, we detail the training process for DiffVAS. The proposed DiffVAS framework comprises two modules: the conditional generative module (CGM), and the task-conditioned planning module (TCPM). Since each module is trained independently, we discuss the training details for each module separately, beginning with the CGM.

%\smallskip
\paragraph{Details of CGM}
We use Stable Diffusion v2.1~\citep{Rombach_2022_CVPR} as CGM's primary latent diffusion model. We integrate partially observed glimpses into the diffusion model by attaching a trainable adapter module following~\citep{zhang2023adding}. The diffusion model\footnote{https://huggingface.co/stabilityai/stable-diffusion-2-1} is kept frozen while the adapter module is fully optimized end-to-end. We use an empty string as the input prompt to the latent diffusion model during training and inference. We randomly mask the ground truth image during training and pass it to the adapter module. We use the Adam optimizer~\citep{kingma2014adam} and a learning rate of 1e-5 to optimize the adapter module. % The adapter module is trained for approximately 100 GPU hours. 
%Finally, the planner module consists of an actor and a critic network. Both these networks are simple MLPs, each comprising three hidden layers with Tanh non-linear activation layers in between. We also incorporate a softmax activation at the final layer of the actor-network to output a probability distribution over the actions. We use a learning rate of 1e-4, batch size of 1, number of training epochs as 300, and the Adam optimizer to train both the actor and critic network. We use the loss function as defined in equation~\ref{eq:ppo} to optimize both the actor and critic network. We choose the values of $\alpha$ and $\beta$ (as defined in equation~\ref{eq:ppo}) to be 0.5 and 0.01 respectively. We also choose the clipping ratio ($\epsilon$) to be 0.2. We select discount factor $\gamma$ to be 0.99 for all the experiments and copy the parameters of $\pi$ onto $\pi_{old}$ after every 4 epochs of policy training. Note that in this work we consider an action invalid if it causes an agent to move outside the predefined search area. During training, we divide the images into $5 \times 5$ non-overlapping pixel grids each of size 300 × 300.

%\smallskip
\paragraph{Details of TCPM}
The planner module consists of four main components: (1) a CLIP text encoder~\citep{radford2021learning}; (2) a cross-attention module; (3) a positional encoding module; and (4) an actor-critic network with a shared backbone. The CLIP text encoder provides information about the target category to the planner module, by computing the text embedding of the target category. The cross-attention module consists of a single cross-attention block~\footnote{https://style-aligned-gen.github.io/} that fuses the information from the CGM and the CLIP text encoder. The positional encoding module provides TCPM with two key pieces of information: (i) the relative positions of each patch in the latent representation space; and (ii) the positions of the revealed patches, and the position of the patches that CGM has reconstructed. Finally, the actor and critic network consists of a lightweight shared backbone comprising 3 convolutional filters and pooling operations. The actor head and critic head are simple MLPs, each comprising three hidden layers with Tanh non-linear activation layers in between. We also incorporate a softmax activation at the final layer of the actor network to output a probability distribution over the grid cells. Except for the CLIP text encoder~\footnote{https://huggingface.co/openai/clip-vit-base-patch32}, all the components of the TCPM are trainable. We use a learning rate of 1e-4, batch size of 1, number of training steps as 100,000, and the Adam optimizer. The script for training and inference of DiffVAS can be accessed through the link provided  \href{https://github.com/mvrl/Multi-objective-active-sampling-for-inpainting}{\textcolor{blue}{here}}.
%We use the loss function as defined in equation~\ref{eq:ppo} to optimize both the actor and critic network. We choose the values of $\alpha$ and $\beta$ (as defined in equation~\ref{eq:ppo}) to be 0.5 and 0.01 respectively. We also choose the clipping ratio ($\epsilon$) to be 0.2. We select discount factor $\gamma$ to be 0.99 for all the experiments and copy the parameters of $\pi$ onto $\pi_{old}$ after every 4 epochs of policy training. Note that in this work we consider an action invalid if it causes an agent to move outside the predefined search area. During training, we divide the images into $5 \times 5$ non-overlapping pixel grids each of size 300 × 300.

%\smallskip
\paragraph{Compute resources}
We use a single NVidia H100 GPU server with a memory of 80 GB for training and a single NVidia V100 GPU server with a memory of 32 GB for running the inference. It requires approximately 50 GPU hours to train TCPM, while the adapter module is optimized for approximately 100 GPU hours. The inference time is 22 seconds for a single search task on a single NVidia V100 GPU, with a maximum exploration budget $\mathcal{B}$ of 10.
Precisely, our end-to-end DiffVAS framework infers the next region to query in approximately 2.20 seconds on a standard NVIDIA V100 GPU. Specifically, a diffusion-based CGM module approximately takes 1.07 seconds to reconstruct a single full image on NVIDIA V100 GPU with 32GB of GPU memory. Given that verifying a search query by a park ranger typically takes a few minutes to hours (depending on the search space), the response time of our system is well within the operational requirements. This efficiency ensures that our framework can provide timely and actionable support, making it highly suitable for real-world deployment where swift decision-making is crucial.

\subsection{Visualization of Exploration behavior of DiffVAS}\label{app:exp_vis}
\begin{figure*}
%\vspace{-4mm}
  \centering
  \setlength{\tabcolsep}{1pt}
  \renewcommand{\arraystretch}{1}
  \includegraphics[width=1.0\linewidth]{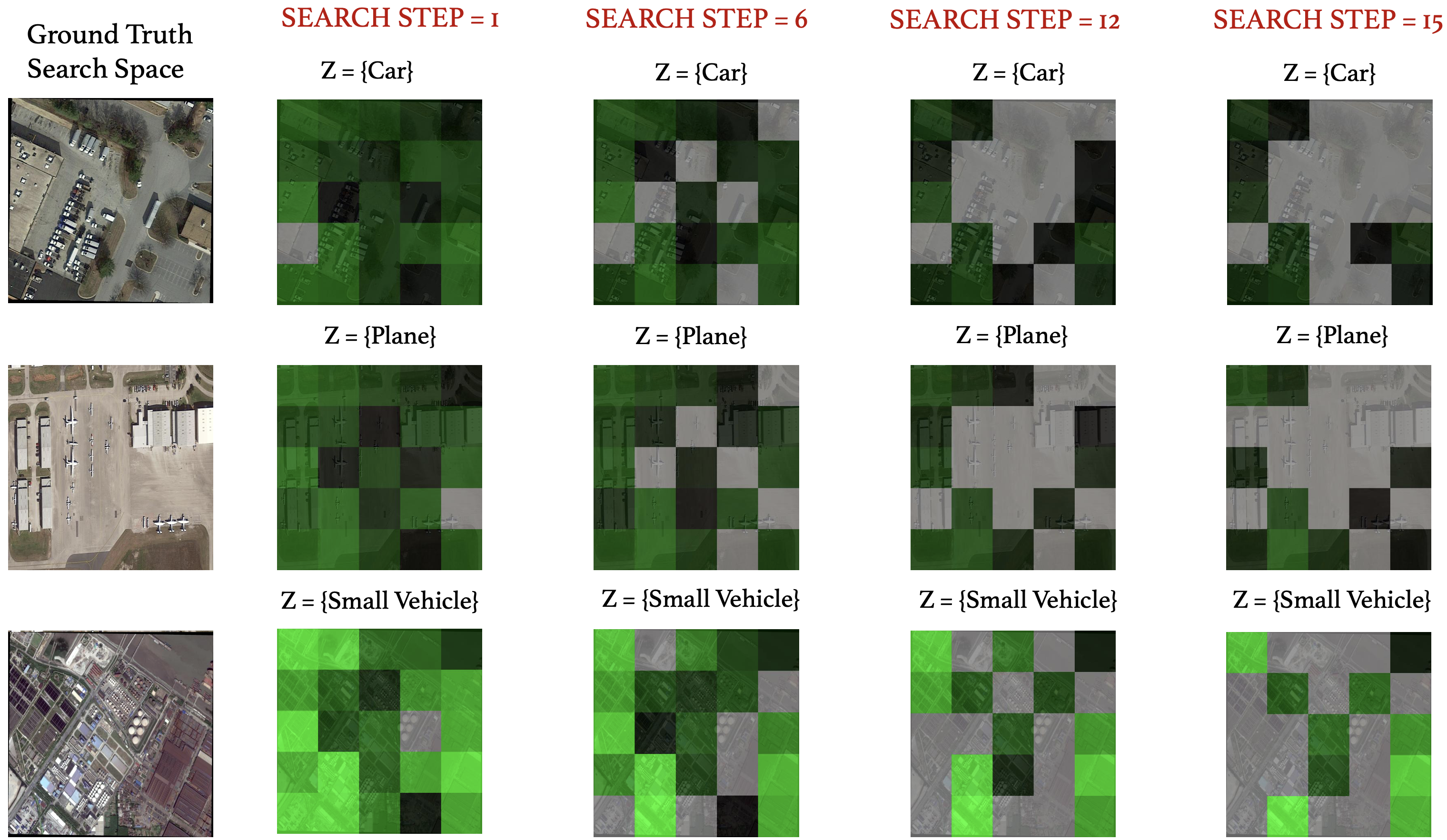}
  \caption{Query sequences for different target category sets $\mathcal{Z}$, as well as corresponding heat maps (darker indicates higher probability). Note that as the search proceeds, the agent becomes relatively more confident (lower entropy) in terms of where it wants to query next.}% under uniform cost.}
  \label{fig:vis_tc-povas_supp3}
%\vspace{-3mm}
\end{figure*}

\begin{figure*}
%\vspace{-4mm}
  \centering
  \setlength{\tabcolsep}{1pt}
  \renewcommand{\arraystretch}{1}
  \includegraphics[width=1.0\linewidth]{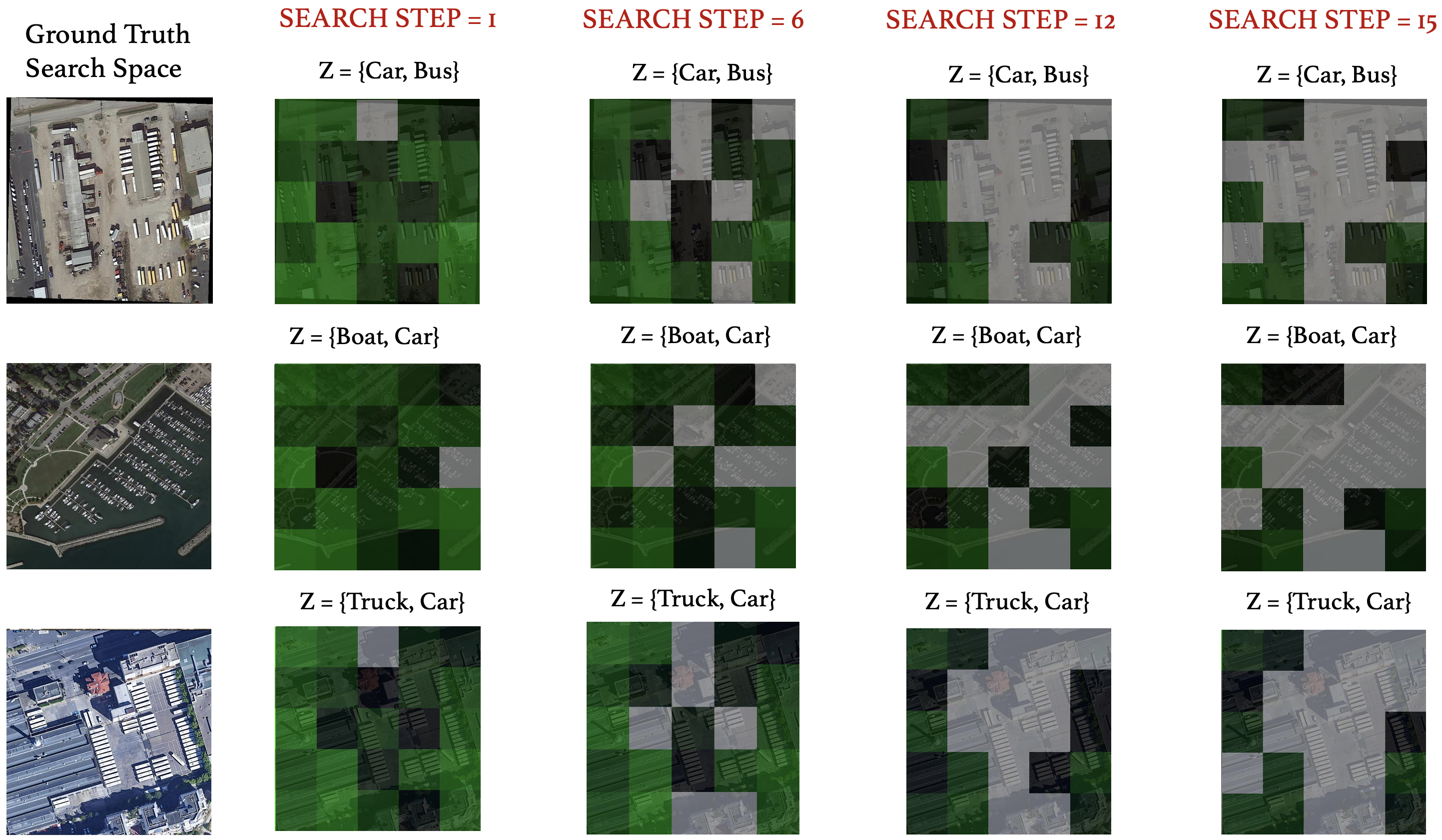}
  \caption{Query sequences for different target category sets $\mathcal{Z}$, as well as corresponding heat maps (darker indicates higher probability). Note that as the search proceeds, the agent becomes relatively more confident (lower entropy) in terms of where it wants to query next.}% under uniform cost.}
  \label{fig:vis_tc-povas_supp2}
%\vspace{-3mm}
\end{figure*}

\begin{figure*}
%\vspace{-4mm}
  \centering
  \setlength{\tabcolsep}{1pt}
  \renewcommand{\arraystretch}{1}
  \includegraphics[width=1.0\linewidth]{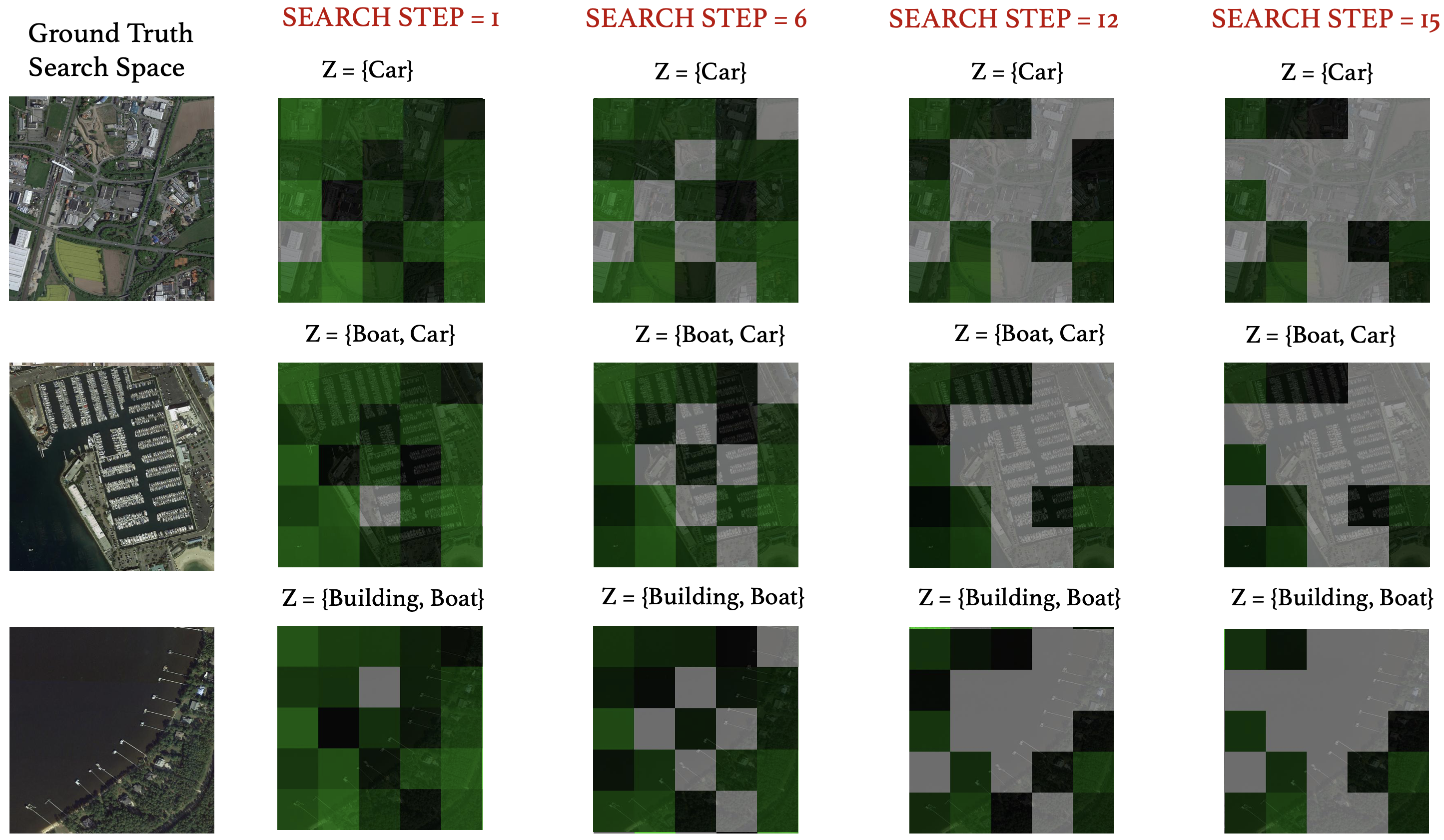}
  \caption{Query sequences for different target category sets $\mathcal{Z}$, as well as corresponding heat maps (darker indicates higher probability). Note that as the search proceeds, the agent becomes relatively more confident (lower entropy) in terms of where it wants to query next.}% under uniform cost.}
  \label{fig:vis_tc-povas_supp1}
%\vspace{-3mm}
\end{figure*}

In this section, we showcase visualizations of exploration behaviors of DiffVAS across diverse search tasks, covering both single- and multi-target category searches. These visualizations are obtained using the DiffVAS policy trained with the DOTA dataset. We depict the visualizations in Fig.~\ref{fig:vis_tc-povas_supp3}, \ref{fig:vis_tc-povas_supp2} and \ref{fig:vis_tc-povas_supp1}. These visualizations provide a comprehensive view of the exploration behaviors of DiffVAS, enabling a nuanced comparison of how the policy adapts to different target specifications.

\subsection{Effectiveness of Cross-attention Layer}\label{app:cross_att}
Here, we analyze the efficacy of the cross-attention layer in learning a target-aware representation suitable for planning. To evaluate the role of the cross-attention layer in the target-conditioned planner module, we remove it and instead concatenate $l_{\mathrm{img}}(t)$ and $l_z$ channel-wise to derive the target-aware representation $l_{\mathrm{img}}^z(t)$, while keeping the rest of the DiffVAS framework unchanged. We refer to the resulting framework as \emph{Concat-DiffVAS}. We compare the performance of DiffVAS and Concat-DiffVAS across different target category sets $\mathcal{Z}$ using the DOTA dataset, as shown in Table~\ref{tab: attention}. We see that in most cases, Concat-DiffVAS obtains a lower ANT score, which indicates that the cross-attention layer is effective in learning target-aware representations for planning.
\begin{table}[H]
%\vspace{-3mm}
    \centering
    %\small
    %\footnotesize
    \small
    \caption{Effectiveness of cross-attention layer in learning a target-aware representation.}
    %\vspace{-3mm}
    \setlength{\tabcolsep}{2.2pt} % Adjust column spacing for better fit
    \begin{tabular}{p{1.87cm}p{0.73cm}p{0.73cm}p{0.86cm}p{0.73cm}p{0.73cm}p{0.86cm}p{0.73cm}p{0.73cm}p{0.86cm}}
        \toprule
        \multicolumn{4}{c}{\>\>\>Test with $\mathcal{Z}$ = \{ Large Vehicle \}} & \multicolumn{3}{c}{Test with $\mathcal{Z}$ = \{ Helicopter \}} & \multicolumn{3}{c}{Test with $\mathcal{Z}$ = \{ Plane \}}\\
        \midrule
        Method & $\mathcal{B}=5$ & $\mathcal{B}=7$ & $\mathcal{B}=10$ & $\mathcal{B}=5$ & $\mathcal{B}=7$ & $\mathcal{B}=10$ & $\mathcal{B}=5$ & $\mathcal{B}=7$ & $\mathcal{B}=10$ \\
        %\midrule
        \cmidrule(r){1-4} \cmidrule(r){5-7} \cmidrule(l){8-10} % For better column separation, thanks to Mico!
        Concat-DiffVAS & 2.02 & 3.13  & \textbf{4.21} & 2.36 & 3.41 & \textbf{5.94} & 3.04 & 3.92 & 5.12 \\
        \hline 
        \textbf{DiffVAS} & \textbf{2.12} & \textbf{3.22} & 3.91 & \textbf{2.54} & \textbf{3.57} & 5.78 & \textbf{3.12} & \textbf{4.07} & \textbf{5.24} \\ 
        \hline
        \hline
        \multicolumn{4}{c}{Test with $\mathcal{Z}$ = \{ Roundabout \}} & \multicolumn{3}{c}{Test with $\mathcal{Z}$ = \{ Ship \}} & \multicolumn{3}{c}{Test with $\mathcal{Z}$ = \{ Harbor \}}\\
        \midrule
        Method & $\mathcal{B}=5$ & $\mathcal{B}=7$ & $\mathcal{B}=10$ & $\mathcal{B}=5$ & $\mathcal{B}=7$ & $\mathcal{B}=10$ & $\mathcal{B}=5$ & $\mathcal{B}=7$ & $\mathcal{B}=10$ \\
        %\midrule
        \cmidrule(r){1-4} \cmidrule(l){5-7} \cmidrule(l){8-10} % For better column separation, thanks to Mico!
        Concat-DiffVAS & 1.91 & 3.08 & 4.42 & 2.30 & 3.27 & \textbf{4.95} & 2.04 & 3.80 & \textbf{5.11} \\
        \hline
        \textbf{DiffVAS} & \textbf{2.01} & \textbf{3.15} & \textbf{4.45} & \textbf{2.32} & \textbf{3.33} & 4.89 & \textbf{2.12} & \textbf{3.91} & 5.05 \\ 
        \bottomrule
    \end{tabular}
    \label{tab: attention}
    %\vspace{-4mm}
\end{table}
\subsection{Comparison with fully observable search setting}\label{app:full}
To evaluate the efficacy of the proposed DiffVAS framework, we compare its performance to a similar approach that assumes full observability of the search space, referred to as \emph{FullVAS}. FullVAS is identical to DiffVAS, except it provides the full search space image $x$ to the $e^{\mathrm{CGM}}$ feature extractor to derive the latent representation of search space (denoted as $l_{\mathrm{img}}$), i.e., $l_{\mathrm{img}}(t) = e^{\mathrm{CGM}}(x)$.
Interestingly, we see from Table~\ref{tab: FULL_OBS} that DiffVAS achieves results comparable to those of FullVAS, despite the fact that DiffVAS never observes the full search area as FullVAS does. This further showcases the strength of our proposed approach, and highlights the strong benefit of the diffusion-based CGM module which reconstructs the underlying search area on the fly.  
%We observe a performance increase of FullVAS relative to DiffVAS, ranging from xx to yy$\%$ across various target category settings on the DOTA dataset, as reported in Table~\ref{tab: FULL_OBS}. The empirical results highlight the inherent challenges of learning an efficient search policy in partially observed environments. We hope our findings will encourage researchers to develop improved frameworks that help close the performance gap further.
\begin{table}[H]
%\vspace{-3mm}
    \centering
    %\small
    %\footnotesize
    \small
    \caption{DiffVAS achieves results that are on average very close to those of FullVAS, despite the fact that DiffVAS never observes the entire search area (which FullVAS does).}
    %\vspace{-3mm}
    \setlength{\tabcolsep}{2.2pt} % Adjust column spacing for better fit
    \begin{tabular}{p{1.97cm}p{0.73cm}p{0.73cm}p{0.86cm}p{0.73cm}p{0.73cm}p{0.86cm}p{0.73cm}p{0.73cm}p{0.86cm}}
        \toprule
        \multicolumn{4}{c}{\>\>\>Test with $\mathcal{Z}$ = \{ Large Vehicle \}} & \multicolumn{3}{c}{Test with $\mathcal{Z}$ = \{ Helicopter \}} & \multicolumn{3}{c}{Test with $\mathcal{Z}$ = \{ Plane \}}\\
        \midrule
        Method & $\mathcal{B}=5$ & $\mathcal{B}=7$ & $\mathcal{B}=10$ & $\mathcal{B}=5$ & $\mathcal{B}=7$ & $\mathcal{B}=10$ & $\mathcal{B}=5$ & $\mathcal{B}=7$ & $\mathcal{B}=10$ \\
        %\midrule
        \cmidrule(r){1-4} \cmidrule(r){5-7} \cmidrule(l){8-10} % For better column separation, thanks to Mico!
        FullVAS & \textbf{2.31} & \textbf{3.32} & 3.87& \textbf{2.62} & \textbf{3.76} & \textbf{5.89} & \textbf{3.22} & \textbf{4.19} & 5.19 \\
        \hline 
        \textbf{DiffVAS} & 2.12 & 3.22 & \textbf{3.91} & 2.54 & 3.57 & 5.78 & 3.12 & 4.07 & \textbf{5.24} \\  
        \hline
        \hline
        \multicolumn{4}{c}{Test with $\mathcal{Z}$ = \{ Roundabout \}} & \multicolumn{3}{c}{Test with $\mathcal{Z}$ = \{ Ship \}} & \multicolumn{3}{c}{Test with $\mathcal{Z}$ = \{ Harbor \}}\\
        \midrule
        Method & $\mathcal{B}=5$ & $\mathcal{B}=7$ & $\mathcal{B}=10$ & $\mathcal{B}=5$ & $\mathcal{B}=7$ & $\mathcal{B}=10$ & $\mathcal{B}=5$ & $\mathcal{B}=7$ & $\mathcal{B}=10$ \\
        %\midrule
        \cmidrule(r){1-4} \cmidrule(l){5-7} \cmidrule(l){8-10} % For better column separation, thanks to Mico!
        FullVAS & \textbf{2.20} & \textbf{3.31} & 4.43 & \textbf{2.35} & \textbf{3.41} & \textbf{4.95} & \textbf{2.22} & \textbf{4.03} & \textbf{5.15} \\
        \hline
        \textbf{DiffVAS} & 2.01 & 3.15 & \textbf{4.45} & 2.32 & 3.33 & 4.89 & 2.12 & 3.91 & 5.05 \\  
        \bottomrule
    \end{tabular}
    \label{tab: FULL_OBS}
    %\vspace{-4mm}
\end{table}
%\begin{comment}
\subsection{More Visualizations of CGM Search Space Reconstructions from Partially Observed Glimpses}\label{app:more_vis}
\begin{figure*}
%\vspace{-4mm}
  \centering
  \setlength{\tabcolsep}{1pt}
  \renewcommand{\arraystretch}{1}
  \includegraphics[width=1.0\linewidth]{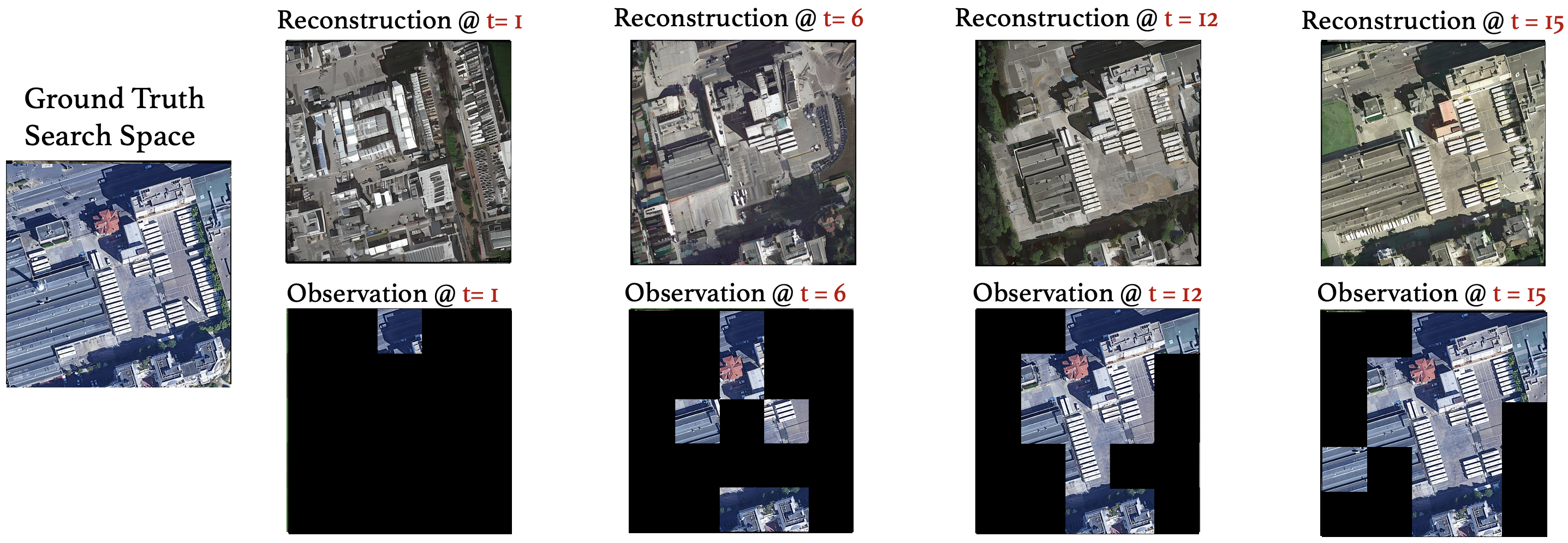}
  \caption{Visualizations of CGM’s reconstruction of the search space from partially observed glimpses at various stages of the search. The reconstruction quality improves as more patches are revealed.}% under uniform cost.}
  \label{fig:more_vis1}
%\vspace{-3mm}
\end{figure*}

\begin{figure*}
%\vspace{-4mm}
  \centering
  \setlength{\tabcolsep}{1pt}
  \renewcommand{\arraystretch}{1}
  \includegraphics[width=1.0\linewidth]{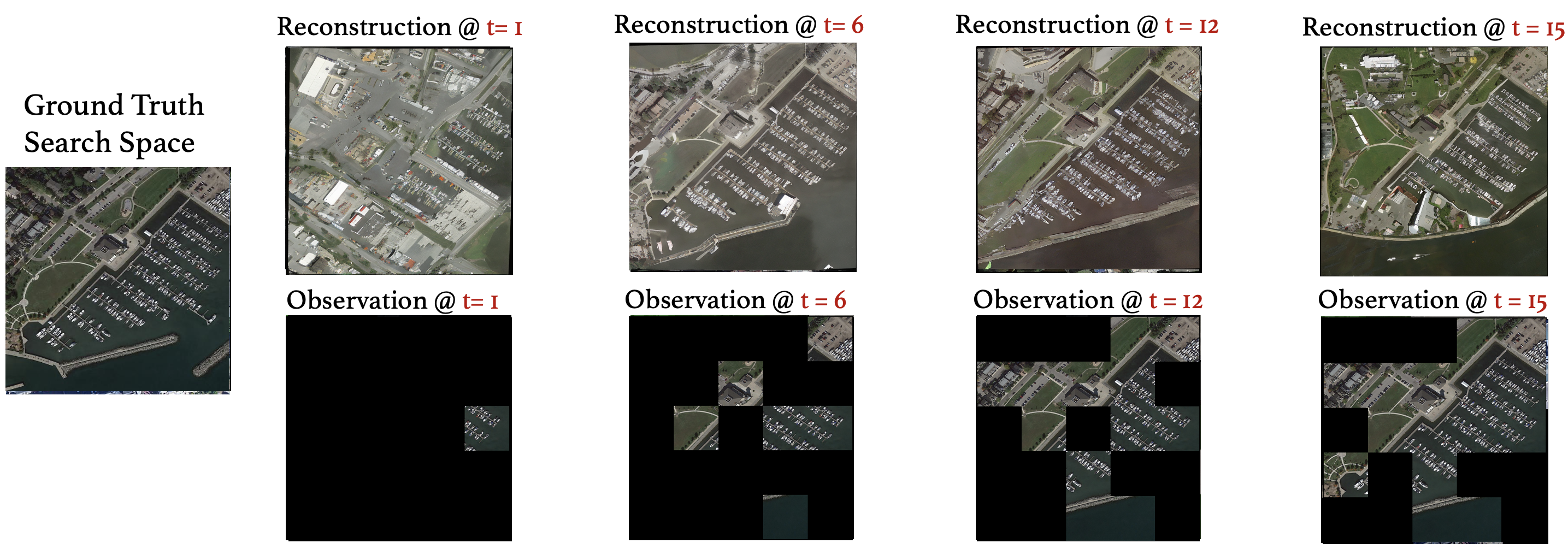}
  \caption{Visualizations of CGM’s reconstruction of the search space from partially observed glimpses at various stages of the search. The reconstruction quality improves as more patches are revealed.}% under uniform cost.}
  \label{fig:more_vis2}
%\vspace{-3mm}
\end{figure*}

\begin{figure*}
%\vspace{-4mm}
  \centering
  \setlength{\tabcolsep}{1pt}
  \renewcommand{\arraystretch}{1}
  \includegraphics[width=1.0\linewidth]{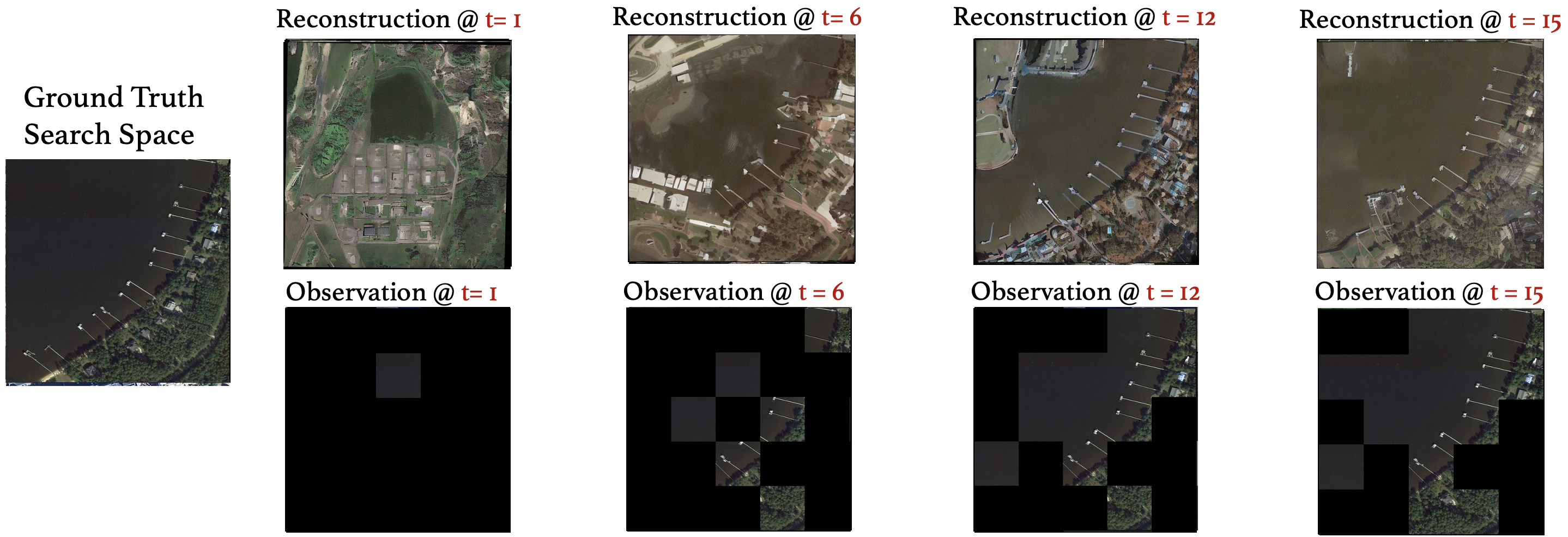}
  \caption{Visualizations of CGM’s reconstruction of the search space from partially observed glimpses at various stages of the search. The reconstruction quality improves as more patches are revealed.}% under uniform cost.}
  \label{fig:more_vis3}
%\vspace{-3mm}
\end{figure*}

In this section, we present additional visualizations of search space reconstruction by CGM from partially observed glimpses at different stages of the search. See figure~\ref{fig:more_vis1},~\ref{fig:more_vis2},~\ref{fig:more_vis3}.

%\end{comment}
\subsection{Reasoning for Selecting the Target Categories for Evaluation}\label{app:select}

In both xView and DOTA, certain classes are extremely rare within the dataset. For instance, images containing at least one instance of class Cement-Mixer appear only once. Consequently, including such classes as targets is not meaningful, as evaluating performance based on a single image does not provide robust analysis/results. Therefore, we excluded these classes. A similar situation applies to multi-target categories. This consideration forms the primary motivation behind our selection of target classes. Note that for zero-shot evaluation on xView, we train models on DOTA using categories that do not appear in xView.

\subsection{Procedure for Sampling an Episode During Training}\label{app:sampling}

During training, we randomly sample an image from the training set and select a random budget within the range \{1 to N-1\} (where N is the number of grids). We then extract the set of target categories if at least one instance of each category is present in the image, based on the ground truth annotations (which also include the precise locations of these target objects as a bounding box). Once the target category set is determined, we randomly select a category from this set to train our policy. The rationale for using a random budget and random target is to ensure that the policy learns to be both target and budget-agnostic. Across various experiments, we demonstrate that DiffVAS consistently outperforms the baseline across different budgets and target categories.

\subsection{Future Directions and Extensions of TC-POVAS Task Setup}\label{app:FD}
Our work represents a foundational step in addressing visual active search (VAS) in partially observable environments, an emerging area with significant real-world implications. While we establish a strong basis that allows for reproducible and controllable experimentation, there are several promising avenues for further exploration and extension.

One key direction is \emph{multi-scale VAS}, which would allow the search agent to operate across different altitudes or resolutions, making the approach even more applicable to real-world scenarios. While our current framework assumes a fixed altitude, real-world settings may require dynamic altitude adjustments to optimize visibility and search efficiency. However, there are also practical constraints where maintaining a constant altitude is necessary, such as UAV operations in hurricane-hit areas where strong winds at higher altitudes could compromise stability. Future work could explore adaptive multi-scale search strategies that balance these trade-offs. 

Another important extension involves enhancing \emph{DiffVAS for more challenging detection scenarios}. While our approach has been rigorously evaluated across diverse object categories and environments, integrating an off-the-shelf object detection module could further improve its resilience against occlusions, noise, and other real-world challenges. This would enable a more robust deployment in dynamic, unpredictable settings where objects may be partially hidden or affected by environmental factors. While our current task formulation provides a structured and meaningful search challenge, alternative formulations could introduce different trade-offs and insights. For example, a variant where the agent observes all patches between its previous and current location might simplify exploration but could lead to trivial search strategies, where the optimal policy would always be to move to the furthest uncovered location to maximize the area revealed. Additionally, non-horizontal and non-vertical movements would introduce ambiguity in determining which intermediate areas to reveal. A possible refinement in future work could involve incorporating blurred observations of these intermediate areas, reflecting real-world movement constraints.

Lastly, \emph{transitioning DiffVAS from simulation to real-world deployment} presents exciting opportunities and challenges, which are, however, beyond the scope of this paper. Key steps include (i) adapting models trained on emulated satellite data to drone imagery, (ii) handling real-world limitations such as imperfect actuation and motion-induced blur, and (iii) optimizing the approach for on-device execution with limited computational resources. Leveraging recent advancements in diffusion models -- such as techniques that reduce reverse diffusion steps -- could make real-time or near-real-time execution more feasible.

%While our current study serves as a foundational contribution, these future directions could drive further advancements, bringing Active Search in Partially Observable Environments closer to real-world applications and bridging the gap between research and practical deployment.
While our current work lays the groundwork for multi-target active search in partially observable environments, these future directions can push the field forward, bridging the gap between research and real-world deployment.

\subsection{Further Details of Cross-Attention Layer}\label{app:cross}

We depict the cross-attention layer in Figure~\ref{fig:framework}. Note that, by \enquote{reference features}, we refer to the latent features extracted by the CGM module, specifically the latent representation of the reconstructed image (i.e., $l_{img}(t)$). On the other hand, \enquote{target features} refer to the latent features of the target category, which are computed using the CLIP-based text encoder (i.e.,$l_{z}$). Additionally, \enquote{AdaIN} stands for \enquote{adaptive instance normalization}, as originally proposed in~\cite{huang2017arbitrary}.

\subsection{Impact Statement}\label{app:is}
In this work we have introduced \emph{DiffVAS}, which is a versatile and adaptive framework for visual active search (VAS) in partially observable environments, making it a valuable tool for real-world applications requiring strategic exploration under uncertainty. By leveraging diffusion-based scene reconstruction and reinforcement learning-driven planning, DiffVAS can efficiently search for multiple target categories simultaneously, a crucial advancement for tasks such as search-and-rescue operations, environmental monitoring, and anti-poaching surveillance. Its ability to reconstruct unseen regions from limited observations enables cost-effective and time-sensitive decision-making, enhancing deployment potential in disaster response, law enforcement, and humanitarian aid.

However, the responsible use of DiffVAS is essential. While it is designed for ethical and beneficial applications, misuse in privacy-invasive surveillance or unauthorized tracking must be strictly avoided. Ensuring that this technology aligns with ethical guidelines and legal frameworks is paramount to maximizing its positive societal impact.

%%%%%%%%%%%%%%%%%%%%%%%%%%%%%%%%%%%%%%%%%%%%%%%%%%%%%%%%%%%%%%%%%%%%%%%%

\end{document}